\documentclass[a4paper,fleqn]{cas-dc}
\usepackage{amsfonts}
\usepackage{amssymb}
\usepackage{amsmath,upgreek}
\usepackage{amsfonts,amssymb}
\usepackage{multicol}
\usepackage{subcaption}
\usepackage{algorithm}
\usepackage{algorithmic}
\usepackage{booktabs}
\usepackage{multirow}
\usepackage{threeparttable}
\usepackage{bm}
\usepackage{xcolor}
\newcommand{\first}[1]{\textbf{\textcolor[RGB]{192,0,0}{#1}}}
\newcommand{\second}[1]{\underline{\textbf{\textcolor[RGB]{112,48,160}{#1}}}}
\usepackage{newfloat}
\usepackage{listings}
\DeclareCaptionStyle{ruled}{labelfont=normalfont,labelsep=colon,strut=off} % DO NOT CHANGE THIS
\floatstyle{ruled}
\newfloat{listing}{tb}{lst}{}
\floatname{listing}{Listing}
\usepackage[numbers]{natbib}
\def\tsc#1{\csdef{#1}{\textsc{\lowercase{#1}}\xspace}}
\tsc{WGM}
\tsc{QE}
\begin{document}
\let\WriteBookmarks\relax
\def\floatpagepagefraction{1}
\def\textpagefraction{.001}
\let\printorcid\relax % 可去掉页面下方的ORCID(s)

% \shorttitle{<short title of the paper for running head>} 
\shorttitle{Is Mamba Effective for Time Series Forecasting?}    

% \shortauthors{<short author list for running head>}
\shortauthors{Zihan Wang et al.}

\title[mode = title]{Is Mamba Effective for Time Series Forecasting?}  

\author{Zihan Wang}
\ead{2310744@stu.neu.edu.cn} 

\author{Fanheng Kong}
\ead{kongfanheng@stumail.neu.edu.cn}

\author{Shi Feng}
\cormark[1]
\ead{fengshi@cse.neu.edu.cn}

\author{Ming Wang}
\ead{sci.m.wang@gmail.com}

\author{Xiaocui Yang}
\ead{yangxiaocui@stumail.neu.edu.cn}

\author{Han Zhao}
\ead{2272065@stu.neu.edu.cn}

\author{Daling Wang}
\ead{wangdaling@cse.neu.edu.cn}

\author{Yifei Zhang}
\ead{zhangyifei@cse.neu.edu.cn}

\address{Department of Computer Science and Engineering, Northeastern University, Shenyang, China}

\cortext[1]{Corresponding author} 

\begin{abstract}
In the realm of time series forecasting (TSF), it is imperative for models to adeptly discern and distill hidden patterns within historical time series data to forecast future states. 
Transformer-based models exhibit formidable efficacy in TSF, primarily attributed to their advantage in apprehending these patterns. 
However, the quadratic complexity of the Transformer leads to low computational efficiency and high costs, which somewhat hinders the deployment of the TSF model in real-world scenarios. 
Recently, Mamba, a selective state space model, has gained traction due to its ability to process dependencies in sequences while maintaining near-linear complexity. 
For TSF tasks, these characteristics enable Mamba to comprehend hidden patterns as the Transformer and reduce computational overhead compared to the Transformer. 
Therefore, we propose a Mamba-based model named Simple-Mamba (S-Mamba) for TSF. 
Specifically, we tokenize the time points of each variate autonomously via a linear layer. 
A bidirectional Mamba layer is utilized to extract inter-variate correlations and a Feed-Forward Network is set to learn temporal dependencies. 
Finally, the generation of forecast outcomes through a linear mapping layer. 
Experiments on thirteen public datasets prove that S-Mamba maintains low computational overhead and achieves leading performance. 
Furthermore, we conduct extensive experiments to explore Mamba's potential in TSF tasks. 
Our code is available at \url{https://github.com/wzhwzhwzh0921/S-D-Mamba}. 
\end{abstract}


% \begin{highlights}
% \item We implement the latest state space model, Mamba, to propose a novel model Simple-Mamba (S-Mamba) for Time Series Forecasting. It utilizes a bidirectional Mamba for extracting inter-variate correlations and assigns the task of processing temporal dependencies to a Feed-Forward Network. 
% \item We conduct comparative evaluations of S-Mamba with nine representative and state-of-the-art models across thirteen public datasets revealing that S-Mamba not only excels in performance but also ensures efficiency with reduced overhead. 
% \item Investigating Mamba’s potential within the TSF field, we conduct a series of detailed experiments. These experiments demonstrate Mamba's superior capabilities in processing TSF data, even indicating a promising potential to displace Transformer in this sector.
% \end{highlights}

% Keywords
\begin{keywords}
Time Series Forecasting \sep 
State Space Model \sep 
Mamba \sep
Transformer
\end{keywords}

\maketitle

% Main text
\section{Introduction}
\begin{figure}[h]
 \centering
 \includegraphics[width=0.35\textwidth]{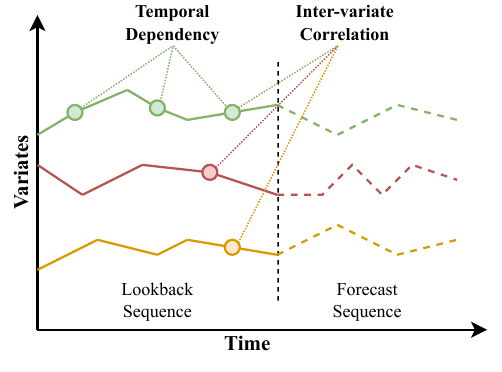}
 \caption{An example of Time Series Forecasting. Lines of different colors represent different variates, with solid lines indicating the historical changes of variates, and dotted lines indicating the future changes that need to be forecasted.}
 \label{fig:TSF}
 \vspace{-5mm}
\end{figure}

Time series forecasting (TSF) involves leveraging historical information from lookback sequence to forecast states in the future \citep{de200625} as Fig. \ref{fig:TSF}. 
These data often have built-in patterns including the temporal dependency (TD), e.g. morning and evening peak patterns in traffic forecast tasks, and the inter-variate correlations (VC), e.g. temperature and humidity correlation patterns in weather forecast tasks. 
Discerning and distilling these patterns from time series data can bring better forecasting \citep{benidis_deep_2023}. 
Transformer \citep{vaswani2017attention} exhibits formidable efficacy in TSF, primarily attributed to their inherent advantages in apprehending TD and VC. 
Numerous Transformer-based models with impressive capabilities have been introduced \citep{wu2021autoformer, zhou2022fedformer}, yet the Transformer architecture faces distinct challenges. 
Foremost is its quadratic computational complexity, which leads to a dramatic increase in calculation overhead as the number of variates or the lookback length increases. 
It hinders the deployment of Transformer-based models in real-world TSF scenarios that require the processing of large amounts of data simultaneously or have high real-time requirements. 
Many models attempt to reduce the computational complexity of the transformer in TSF by modifying its structure, such as focusing only on a portion of the sequence \citep{kitaev2020reformer, zhou2021informer, li2019enhancing}. 
The loss of information in the above models may also lead to certain performance degradations. 
A more promising approach involves using linear models instead of transformer \citep{li2023revisiting, zeng2023dlinear}, which possesses linear computational complexity. 
However, linear models relying solely on linear numerical calculations do not incorporate in-context information and are suboptimal compared to state-of-the-art Transformer models. 
And accurate forecasts can only be achieved when sufficient input information is available \citep{zeng2023dlinear}.

The State Space Models (SSM) \citep{gu2021efficiently, smith2022simplified} 
demonstrate potential in simultaneously optimizing performance and computational complexity. 
SSMs employ convolutional calculation to capture sequence information and eliminate hidden states making it benefit from parallel computing and achieving near-linear complexity in processing speed. 
\citet{rangapuram2018deep} attempts to employ SSM for TSF, but the SSM architecture it used is unable to identify and filter content effectively, and the captured dependencies are solely based on distance, resulting in unsatisfactory performance. 
Mamba \citep{gu2023mamba}, introduces a selective mechanism into SSM, enabling it to discern valuable information like the attention mechanism. 
Numerous researchers develop models based on Mamba \citep{zhu2024vision, yang2024vivim}, demonstrating its considerable potential across both text and image domains. 
These Mamba-based models achieve a synergistic balance between enhanced performance and computational efficiency. 
Consequently, we are motivated to explore further the potential of Mamba in TSF. 

We launch a Mamba-based model Simple-Mamba (S-Mamba) for TSF tasks. 
In the S-Mamba, the time points of each variate are tokenized by a linear layer. 
Subsequently, a Mamba VC (Inter-variate Correlation) Encoding layer encodes the VC by utilizing a bidirectional Mamba to leverage the global inter-variate mutual information. 
A Feed-Forward Network (FFN) TD (Temporal Dependency) Encoding Layer containing a simple FFN is followed to extract the TD. 
Ultimately, a mapping layer is utilized to output the forecast results.
Experimental results on thirteen public datasets from traffic, electricity, weather, finance, and energy domains demonstrate that S-Mamba not only has low requirements in GPU memory usage and training time but also maintains superior performance compared to the state-of-the-art models in TSF. 
Concurrently, extensive experiments are conducted to assess the efficacy and potential of Mamba in TSF tasks. 
For instance, we evaluate whether Mamba demonstrates generalization capabilities comparable to those of the Transformer in handling TSF data. 
Our contributions are summarized as follows:
\begin{itemize}
    \item We propose S-Mamba, a Mamba-based model for TSF, which delegates the extraction of inter-variate correlations and temporal dependencies to a bidirectional Mamba block and a Feed-Forward Network. 
    \item We compare the performance of the S-Mamba against representative and state-of-the-art models in TSF. The results confirm that S-Mamba not only delivers superior forecast performance but also requires less computational resources. 
    \item We conduct extensive experiments mainly focusing on exploring the characteristics of Mamba when facing TSF data to further discuss the potential of Mamba in TSF tasks. 
\end{itemize}

\section{Related Work}
\label{sec:related_works}
In conjunction with our work, two main areas of related work are investigated: (1) time series forecasting, and (2) applications of Mamba.

\subsection{Time Series Forecasting}
There have been two main architectures for TSF approaches, which are Transformer-based models \citep{lim2021time,Midilli2023ARF,Zeng2022AreTE} and linear models \citep{benidis_deep_2023,Mahmoud2021,SEZER2020106181}.

\subsubsection{Transformer-based Models}

Transformers are primarily designed for tasks that involve processing and generating sequences of tokens \citep{vaswani2017attention}. The excellent performance of Transformer-based models has also attracted numerous researchers to focus on time series forecasting tasks \citep{ahmed_transformers_2023}. The transformer is utilized by \citet{duong-trung_temporal_2023} to solve the persistent challenge of long multi-horizon time series forecasting. Time Absolute Position Encoding (tAPE) and Efficient implementation of Relative Position Encoding (eRPE) are proposed in \citep{foumani_improving_2024} to solve the position encoding problem encountered by Transformer in multivariate time series classification (MTSC). \citet{WANG2024111321} replace the standard convolutional layer with an dilated convolutional layer and propose Graphformer to efficiently learn complex temporal patterns and dependencies between multiple variates. Some researchers have also considered the application of Transformer-based time series forecasting models in specific domains, such as piezometric level prediction \citep{mellouli_transformers-based_2022}, forecasting crude oil returns \citep{abdollah_pour_new_2022}, predicting the power generation by solar panels \citep{sherozbek_transformers-based_2023}, etc.

While they excel at capturing long-range dependencies in text, they may not be as effective in modeling sequential patterns. The use of content-based attention in Transformers is not effective in detecting essential temporal dependencies, especially for time-series data with weakening dependencies over time and strong seasonality patterns \citep{woo_etsformer_2022}. Particularly, the predictive capability and robustness of Transformer-based models may decrease rapidly when the input sequence is too long \citep{wen_transformers_2023}. Moreover, the $ O(\text{N}^2)$ time complexity makes Transformer-based models cost more computation and GPU memory resources. In addition, the previously mentioned issue of position encoding is also a challenge that deserves attention.

\subsubsection{Linear Models}

In addition to Transformer-based models, many researchers are keen to perform time series forecasting tasks using linear models \citep{benidis_deep_2023}. \citet{chen_tsmixer_2023} proposed TSMixer with all-MLP architecture to efficiently utilize cross-variate and auxiliary information to improve the performance of time series forecasting. LightTS \citep{zhang_less_2022} is dedicated to solving multivariate time series forecasting problems, and it can efficiently handle very long input series. \citet{WANG2024111463} propose Time Series MLP to improve the efficiency and performance of multivariate time series forecasting. \citet{yi_frequency-domain_2023} explores MLP in the frequency domain for time series forecasting and proposes a novel architecture for FreTS that includes two phases: domain conversion and frequency learning.

Compared to Transformer-based models, MLP-based models are simpler in structure, less complex and more efficient. However, the MLP-based models also suffer from a number of shortcomings. In the case of high volatility and non-periodic, non-stationary patterns, MLP performance relying only on past observed temporal patterns is not satisfactory \citep{chen_tsmixer_2023}. In addition, MLP is worse at capturing global dependencies compared to Transformers \citep{yi_frequency-domain_2023} and need longer input than Transformer-based models. 

\subsection{Applications of Mamba}
As a new architecture, Mamba \citep{gu2023mamba} swiftly attracted the attention of a large number of researchers in Natural Language Processing (NLP), Computer Vision (CV), and other Artificial Intelligence communities.

\subsubsection{Mamba in Natural Language Processing}
\citet{pioro_moe-mamba_2024} and \citet{Anthony2024BlackMambaMO} replaced the Transformer architecture in the Mixture of Experts (MoE) with the Mamba architecture, achieving a complete override of Mamba's and Transformer-MoE's performance. Mamba has demonstrated strong performance in clinical note generation \citep{yang_clinicalmamba_2024}. \citet{jiang2024dualpath} replace Transformers with Mamba and demonstrate that Mamba can achieve match or outperform results on speech separation tasks with fewer parameters than Transformer. Empirical evidence is provided using simple NLP tasks (like translation) that Mamba can be an efficient alternative to Transformer for in-context learning tasks with long input sequences \citep{grazzi2024mamba}.

\subsubsection{Mamba in Computer Vision}
Mamba has been used to solve the long-range dependency problem in biomedical image segmentation tasks \citep{ma_u-mamba_2024}. \citet{cao2024novel} propose a local-enhanced vision Mamba block named LEVM to improve local information perception, achieving state-of-the-art results on multispectral pansharpening and multispectral and hyperspectral image fusion tasks. Fusion-Mamba block \citep{dong2024fusionmamba} is designed to map features from images with different types (such as RGB and IR) into a hidden state space for interaction and enhance the representation consistency of features. \citet{liu2024hsidmamba} utilize the proposed HSIDMamba and Bidirectional Continuous Scanning Mechanism to improve the capture of long-range and local spatial-spectral information and improve denoise performance. In addition, Mamba has also been used in small target detection \citep{chen_mim-istd_2024}, medical image reconstruction\citep{huang2024mambamir} and classification \citep{yue2024medmamba}, hyperspectral image classification \citep{yao2024spectralmamba}, etc. 

\subsubsection{Mamba in Others}
In addition to the two single modalities described, the application of Mamba to multimodal tasks has received a lot of attention. VideoMamba \citep{li_videomamba_2024} achieves efficient long-term modeling using Mamba's linear complexity operator, showing advantages on long video understanding tasks. \citet{zhao2024cobra} extend Mamba to a multi-modal large language model to improve the efficiency of inference, achieving comparable performance to LLaVA \citep{liu2023llava} with only about 43\% of the number of parameters. 

Furthermore, Mamba's sequence modeling capabilities have also received attention from researchers. \citet{Schiff2024CaduceusBE} extend long-range Mamba to a BiMamba component that supports bi-directionality, and to a MambaDNA block as the basis of long-range DNA language models. Mamba has also been shown to be effective on the tasks of predicting sequences of sensor data \citep{bhirangi2024hierarchical} and stock prediction \citep{shi2024mambastock}. Sequence Reordering Mamba \citep{yang2024mambamil} are proposed to exploit the inherent valuable information embedded within the long sequences. \citet{ahamed2024timemachine} propose Mamba-based TimeMachine to capture long-term dependencies in multivariate time series data.

As can be seen from the application of Mamba in these areas, Mamba can effectively reduce the parameter size and improve the efficiency of model inference while achieving similar or outperforming performance. It captures global dependencies better in a lightweight structure and has a better sense of position relationships. In addition, the Mamba architecture is more robust. Furthermore, the performance of Mamba in sequence modelling tasks further inspired us to explore whether Mamba can effectively mitigate the issues faced by Transformer-based models and linear models on TSF tasks.

\section{Preliminaries}
\subsection{Problem Statement}
In time series forecasting tasks, the model receives input as a history sequence $U_{in}=[u_1, u_2, \ldots, u_L] \in {\mathbb {R}^{L \times V}}$ and $u_{n} = [p_1, p_2, \ldots, p_V]$. and then uses the information to predict a future sequence $U_{out}=[u_{L+1}, u_{L+2}, \ldots, u_{L+T}] \in {\mathbb {R}^{T \times V}}$. 
The preceding $L$ and $T$ are referred to as the review window and prediction horizon respectively, representing the lengths of the past and future time windows, while $p$ is a variate and $V$ represents the total number of variates.

\subsection{State Space Models}
State Space Models can represent any cyclical process with latent states. 
By using first-order differential equations to represent the evolution of the system's internal state and another set to describe the relationship between latent states and output sequences, input sequences $x(t)\in\mathbb{R}^D$ can be mapped to output sequences $y(t)\in\mathbb{R}^N$ through latent states $h(t)\in\mathbb{R}^N$ in \eqref{equ:latent}:
\begin{equation}
\centering
\label{equ:latent}
\begin{aligned}
h(t)^{'} &= \textbf{\emph A}h(t)+\textbf{\emph B}x(t),\\
y(t) &= \textbf{\emph C}h(t),
\end{aligned}
\end{equation}
where $\textbf{\emph A}\in{\mathbb{R}^{N \times N}}$ and $\textbf{\emph{B,C}} \in {\mathbb{R}^{N \times D}}$ are learnable matrices. 
Then, the continuous sequence is discretized by a step size $\Delta$, and the discretized SSM model is represented as \eqref{equ:discretize}. 
\begin{equation}
\label{equ:discretize}
\begin{aligned}
h_t&=\overline{\textbf{\emph A}}h_{t-1}+\overline{\textbf{\emph B}}x_t,\\
y_t&=\textbf{\emph C}h_t,
\end{aligned}
\end{equation}
where $\overline{\textbf{\emph A}}={\mathrm{exp}}(\Delta A) $ and $\overline{\textbf{\emph{B}}}=(\Delta \textbf{\emph A})^{-1}({\mathrm{exp}}(\Delta \textbf{\emph A})-I) \cdot \Delta \textbf{\emph B}$. 
Since transitioning from continuous form $(\Delta,\textbf{\emph A},\textbf{\emph B},\textbf{\emph C})$ to discrete form $(\overline{\textbf{\emph A}},\overline{\textbf{\emph B}},\textbf{\emph C})$, the model can be efficiently calculated using a linear recursive approach \citep{gu2021CombiningRC}. 
The structured state space model (S4) \citep{gu2021efficiently}, originating from the vanilla SSM, utilizes HiPPO \citep{gu2020HiPPORM} for initialization to add structure to the state matrix $\textbf{\emph A}$, thereby improving long-range dependency modeling. 

\begin{algorithm}[tb]
\caption{The process of Mamba Block}
\label{alg:mamba}
\textbf{Input}: $\bm{X}:(B,V,D)$ \\
\textbf{Output}: $\bm{Y}:(B,V,D)$
\begin{algorithmic}[1] %[1] enables line numbers
\STATE $x,z:(B,V,ED) \leftarrow \mathrm{Linear}(\bm{U})$  \COMMENT{Linear projection}
\STATE $x^{'}:(B,V,ED) \leftarrow \mathrm{SiLU} (\mathrm{Conv1D}(x))$
\STATE $\textbf{\emph A}:(D,N) \leftarrow Parameter$ \COMMENT{Structured state matrix}
\STATE $\textbf{\emph{B,C}}:(B,V,N) \leftarrow \mathrm{Linear}(x^{'}), \mathrm{Linear} (x^{'})$
\STATE $\Delta:(B,V,D) \leftarrow \mathrm{Softplus} (Parameter + \mathrm{Broadcast} (\mathrm{Linear}(x^{'})))$
\STATE $\overline{\textbf{\emph A}},\overline{\textbf{\emph B}}:(B.V.D.N) \leftarrow {discretize}(\Delta,\textbf{\emph A},\textbf{\emph B})$ \COMMENT{Input-dependent parameters and discretization}
\STATE $y:(B,V,ED) \leftarrow \mathrm{SelectiveSSM} (\overline{\textbf{\emph A}},\overline{\textbf{\emph B}},\textbf{\emph C})(x^{'})$
\STATE $y^{'}:(B,V,ED) \leftarrow y \otimes \mathrm{SiLU}(z)$
\STATE $\bm{Y}:(B,V,D) \leftarrow \mathrm{Linear}(y^{'})$ \COMMENT{Linear Projection}

\end{algorithmic}
\end{algorithm}

\begin{figure}[t]
 \centering
 \includegraphics[width=0.2\textwidth]{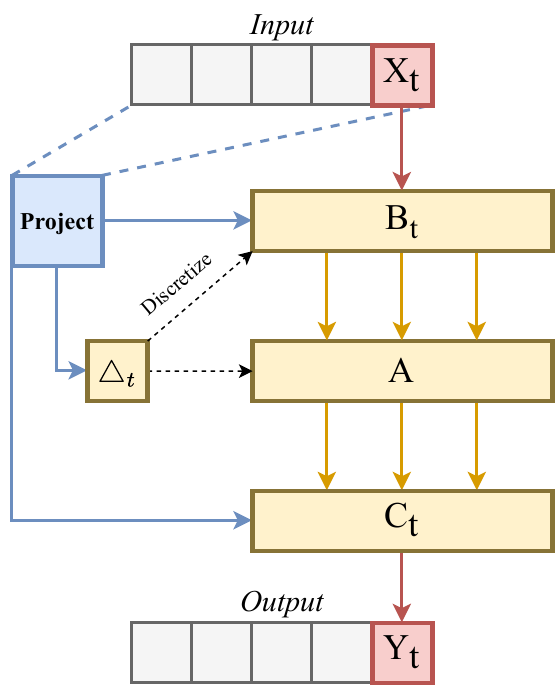}
 \caption{The structure of selective SSM (Mamba).}
 \label{fig:select}
 \vspace{-5mm}
\end{figure}

\subsection{Mamba Block}
Mamba \citep{gu2023mamba} introduces a data-dependent selection mechanism into the S4 and incorporates hardware-aware parallel algorithms in its looping mode. 
The mechanism enables Mamba to capture contextual information in long sequences while maintaining computational efficiency. 
As an approximately linear perplexity series model, Mamba demonstrates potential in long sequence tasks, compared to transformers, in both efficiency enhancement and performance improvement. 
The details are presented in the algorithm related to the mamba layer in Alg.\ref{alg:mamba} and the description in Fig. \ref{fig:select}, where the former illustrates the complete data processing procedure, while the latter depicts the formation process of the output at sequence position $t$. 
The Mamba layer takes a sequence $\bm{X}\in \mathbb{R}^{B \times V \times D}$ as input, where $B$ denotes the batch size, $V$ denotes the number of variates, and $D$ denotes hidden dimension. 

The block first expands the hidden dimension to $ED$ through linear projection, obtaining $x$ and $z$. 
Then, it processes the projection obtained earlier using convolutional functions and a SiLU \citep{Elfwing2017SigmoidWeightedLU} activation function to get $x'$. 
Based on the discretized SSM selected by the input parameters, denoted as the core of the Mamba Block, together with $x'$, it generates the state representation $y$. 
Finally, $y$ is combined with a residual connection from $z$ after activation, and the final output $y_t$ at time step $t$ is obtained through a linear transformation. 
In summary, the Mamba Block effectively handles sequential information by leveraging selective state space models and input-dependent adaptations. 
The parameters involved in the Mamba Block include an SSM state expansion factor $N$, a size of convolutional kernel $k$, and a block expansion factor $E$ for input-output linear projection. 
The larger the values of $N$ and $E$, the higher the computational cost. 
The final output of the Mamba block is $\bm{Y} \in \mathbb{R}^{B \times V \times D}$. 

\begin{figure*}[htp]
    % \vspace{-5mm}
 \centering
 \includegraphics[width=0.85\textwidth]{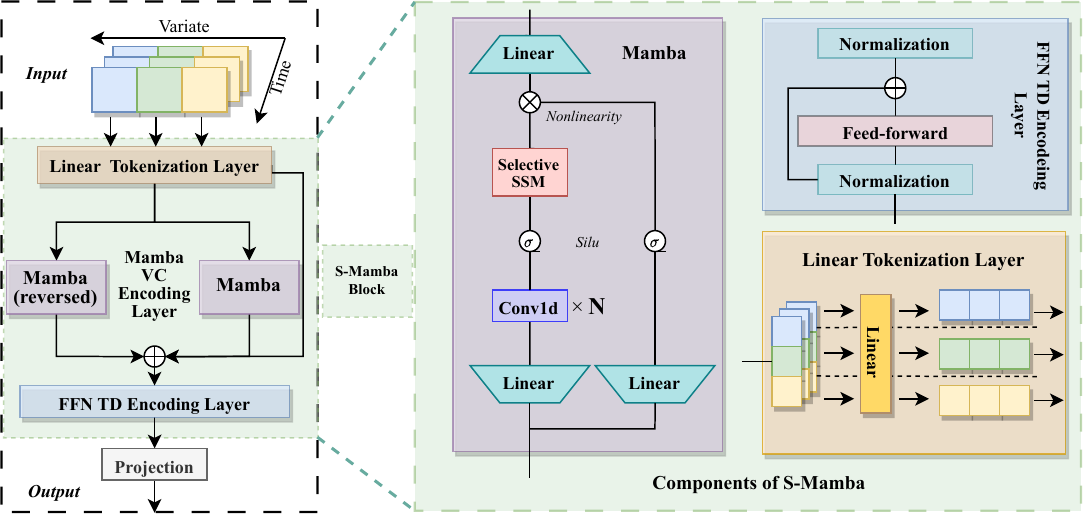}
 \caption{Overall framework of S-Mamba, the left side of the figure presents the overall architecture of our model. The right side of the figure details the components of the S-Mamba Block.}
 \label{fig:MambaTSF}
\end{figure*}

\section{Methodology}
In this section, we provide a detailed introduction of S-Mamba. 
Fig. \ref{fig:MambaTSF} illustrates the overall structure of S-Mamba, which is primarily composed of four layers. 
The first layer, the Linear Tokenization Layer, tokenizes the time series with a linear layer. 
The second layer, the Mamba inter-variate correlation (VC) Encoding layer, employs a bidirectional Mamba block to capture mutual information among variates. 
The third layer, the FFN Temporal Dependencies (TD) Encoding Layer, further learns the temporal sequence information and finally generates future series representations by a Feed-Forward Network.  
Then the final layer, the Projection Layer, is only responsible for mapping the processed information of the above layers as the model's forecast. 
Alg. \ref{alg:S-Mamba} demonstrates the operation process of S-Mamba. 

\begin{algorithm}[tb]
\caption{The Forecasting Procedure of S-Mamba}
\label{alg:S-Mamba}
\textbf{Input}: $Batch(U_{in}) = [u_1,u_2, \ldots u_L] : (B,L,V)$ \\
\textbf{Output}: $Batch(U_{out})= [u_{L+1}, u_{L+2}, \ldots u_{L+T}] : (B,T,V)$
\begin{algorithmic}[1] %[1] enables line numbers
\STATE \textbf{Linear Tokenization Layer:}
\STATE $Batch(U_{in}^\top):(B,V,L) \leftarrow \mathrm{Transpose} (Batch(U_{in}))$ 
\STATE $\bm{U}^{tok}:(B,V,D) \leftarrow \mathrm{Linear Tokenize} (Batch(U_{in}^\top))$ \COMMENT{Tokenization}
\FOR{$l$ in $Mamba~Layers$}
\STATE \textbf{Mamba VC Encoding Layer:}
\STATE $\overrightarrow{\bm{Y}}:(B,V,D) \leftarrow \overrightarrow{\mathrm{Mamba~Block}} (\bm{U})$
\STATE $\overleftarrow{\bm{Y}}:(B,V,D) \leftarrow \overleftarrow{\mathrm{Mamba~Block}} (\bm{U})$
\STATE $\bm{Y}:(B,V,D) \leftarrow \overrightarrow{\bm{Y}} + \overleftarrow{\bm{Y}}$ \COMMENT{Fusion Bidirectional Information}
\STATE $\bm{U}^{'}:(B,V,D) \leftarrow \bm{Y} + \bm{U}$ \COMMENT{Residual Connection}
\STATE \textbf{FFN TD Encoding Layer:}
\STATE $\bm{U}^{'}:(B,V,D) \leftarrow \mathrm{LayerNorm}(\bm{U}^{'})$ 
\STATE $\bm{U}^{'}:(B,V,D) \leftarrow \mathrm{Feed-Forward}(\bm{U}^{'})$ 
\STATE $\bm{U}^{'}:(B,V,D) \leftarrow \mathrm{LayerNorm}(\bm{U}^{'})$
\ENDFOR
\STATE \textbf{Projection:}
\STATE $\bm{U}^{'}:(B,V,T) \leftarrow \mathrm{Projection}(\bm{U}^{'})$
\STATE $Batch(U_{out}):(B,T,V) \leftarrow \mathrm{Transpose}(\bm{U}^{'})$

\end{algorithmic}
\end{algorithm}

\subsection{Linear Tokenization Layer}
The input for the Linear Tokenization Layer is $U_{in}$. 
Similar to iTransformer \citep{liu2023itransformer}, we commence by tokenizing the time series, a method analogous to the tokenization of sequential text in natural language processing, to standardize the temporal series format. 
This pivotal task is executed by a single linear layer in Eq. \eqref{equ:emb}. 
\begin{align}
    \bm{U} = \mathrm{Linear}(Batch(U_{in})), \label{equ:emb}
\end{align}
where $\bm{U}$ is the output of this layer. 
\subsection{Mamba VC Encoding Layer}
Within this layer, the primary objective is to extract the VC by linking variates that exhibit analogous trends aiming to learn the mutual information therein. 
The Transformer architecture confers the capacity for global attention \citep{vaswani2017attention}, enabling the computation of the impact of all other variates upon a given variate, which facilitates the learning of precise information. 
However, the computational load of global attention escalates exponentially with an increase in the number of variates, potentially rendering it impractical. 
This limitation could restrict the application of Transformer-based algorithms in real-world scenarios. 
In contrast, Mamba's selective mechanism can discern the significance of different variates akin to an attention mechanism, and it exhibits a computational overhead that escalates in a near-linear fashion with an increasing count of variates. 
Yet, the unilateral nature of Mamba precludes it from attending to global variates in the manner of the Transformer; its selection mechanism is unidirectional, capable only of incorporating antecedent variates. 
To surmount this limitation, we employ two Mamba blocks to be combined as a bidirectional Mamba layer as Eq. \eqref{equ:mamba2}, which facilitates the acquisition of correlations among all variates. 
\begin{equation}
\begin{aligned}
\label{equ:mamba2}
    \overrightarrow{\bm{Y}} = \overrightarrow{\mathrm{Mamba~Block}} (\bm{U}), \\
    \overleftarrow{\bm{Y}} = \overleftarrow{\mathrm{Mamba~Block}} (\bm{U}).
\end{aligned} 
\end{equation}
The VC encoded by the bidirectional Mamba is aggregated $\bm{Y} = \overrightarrow{\bm{Y}} + \overleftarrow{\bm{Y}}$ and connected with another residual network to form the output of this layer $\bm{U}^{'} = \mathrm{\bm{Y}} + \bm{U}$.

\subsection{FFN TD Encoding Layer}
At this layer, we further process the output of the Mamba VC Encoding Layer.
Firstly, we employ a normalization layer \citep{liu2023itransformer} to enhance convergence and training stability in deep networks by standardizing all variates to a Gaussian distribution, thereby minimizing disparities resulting from inconsistent measurements. 
Then, the feed-forward network (FFN) is used on the series representation of each variate. 
The FFN layer encodes observed time series and decodes future series representations using dense non-linear connections. 
During this procedure, FFN implicitly encodes TD by keeping the sequential relationships. 
Finally, another normalization layer is set to adjust the future series representations. 

\subsection{Projection Layer}
Based on the output of the FFN TD Encoding layer, the tokenized temporal information is reconstructed into the time series requiring prediction via a mapping layer, subsequently undergoing transposition to yield the final predictive outcome. 

\section{Experiments}
\subsection{Datasets and Baselines} 
We conduct experiments on thirteen real-world datasets. 
For convenience of comparison, we divide them into three types.
(1) Traffic-related datasets: Traffic \citep{wu2021autoformer} and PEMS \citep{chen2001freeway}. Traffic is a collection of hourly road occupancy rates from the California Department of Transportation, capturing data from 862 sensors across San Francisco Bay area freeways from January 2015 to December 2016. 
And PEMS is a complicated spatial-temporal time series for public traffic networks in California including four public subsets (PEMS03, PEMS04, PEMS07, PEMS08), which are the same as SCINet \citep{liu2022scinet}. 
Traffic-related datasets are characterized by a large number of variates, most of which are periodic. 
(2) ETT datasets: ETT \citep{zhou2021informer} (Electricity Transformer Temperature) comprises data on load and oil temperature, collected from electricity transformers over the period from July 2016 to July 2018. 
It contains four subsets, ETTm1, ETTm2, ETTh1 and ETTh2. 
ETT datasets have few varieties and weak regularity. 
(3) Other datasets: Electricity \citep{wu2021autoformer}, Exchange \citep{wu2021autoformer}, Weather \citep{wu2021autoformer}, and Solar-Energy \citep{lai2018modeling}. 
Electricity records the hourly electricity consumption of 321 customers from 2012 to 2014. 
Exchange collects daily exchange rates of eight countries from 1990 to 2016. 
Weather contains 21 meteorological indicators collected every 10 minutes from the Weather Station of the Max Planck Biogeochemistry Institute in 2020. 
Solar-Energy contains solar power records in 2006 from 137 PV plants in Alabama State which are sampled every 10 minutes. 
Among them, the Electricity and Solar-Energy datasets contain many variates, and most of them are periodic, while the Exchange and Weather datasets contain fewer variates, and most of them are aperiodic. 
Tab. \ref{tab:datasets} shows the statistics of these datasets.
\begin{table}[htbp]
\begin{center}
\renewcommand{\arraystretch}{0.9}
\caption{The statistics of the thirteen public datasets.}
\label{tab:datasets}
\resizebox{0.45\textwidth}{!}{
    \begin{tabular}{c|ccc}
    \hline
    Datasets & Variates & Timesteps & Granularity  \\
    \hline
    Traffic & 862 & 17,544 & 1hour  \\
    PEMS03 & 358 & 26,209 & 5min  \\
    PEMS04 & 307 & 16,992 & 5min  \\
    PEMS07 & 883 & 28,224 & 5min  \\
    PEMS08 & 170 & 17,856 & 5min  \\ 
    ETTm1 $\&$ ETTm2 & 7 & 17,420 & 15min  \\
    ETTh1 $\&$ ETTh2 & 7 & 69,680 & 1hour  \\
    Electricity & 321 & 26,304 & 1hour  \\
    Exchange & 8 &  7,588 &  1day  \\
    Weather & 21 & 52,696 & 10min  \\
    Solar-Energy & 137 & 52,560 & 10min  \\
    \hline
    \end{tabular}
}
\end{center}
\vspace{-0.7cm}
\end{table} 
\begin{table*}[htbp]
  \caption{Full results of S-Mamba and baselines on traffic-related datasets.
  The lookback length $L$ is set to 96 and the forecast length $T$ is set to 12, 24, 48, 96 for PEMS and 96, 192, 336, 720 for Traffic.}
  \label{tab:results_traffic}
  \renewcommand{\arraystretch}{0.9}
  \centering
  \resizebox{\textwidth}{!}{
  \begin{threeparttable}
  \begin{small}
  \setlength{\tabcolsep}{2.6pt}
  % \vspace{-3mm}
  \begin{tabular}{c|c|cc|cc|cc|cc|cc|cc|cc|cc|cc|cc}
    \toprule
    \multicolumn{2}{c|}{Models} & \multicolumn{2}{c|}{\textbf{S-Mamba}} & \multicolumn{2}{c}{iTransformer} & \multicolumn{2}{c}{RLinear} & \multicolumn{2}{c}{PatchTST} & \multicolumn{2}{c}{Crossformer} & \multicolumn{2}{c}{TiDE} & \multicolumn{2}{c}{TimesNet} & \multicolumn{2}{c}{DLinear} & \multicolumn{2}{c}{FEDformer} &  \multicolumn{2}{c}{Autoformer} \\
    \cmidrule(lr){1-2}\cmidrule(lr){3-4}\cmidrule(lr){5-6}\cmidrule(lr){7-8} \cmidrule(lr){9-10}\cmidrule(lr){11-12}\cmidrule(lr){13-14}\cmidrule(lr){15-16}\cmidrule(lr){17-18}\cmidrule(lr){19-20}\cmidrule(lr){21-22}  
    \multicolumn{2}{c|}{Metric} & MSE & MAE & MSE & MAE & MSE & MAE & MSE & MAE & MSE & MAE & MSE & MAE & MSE & MAE & MSE & MAE & MSE & MAE & MSE & MAE  \\
    \toprule
    \multirow{4}{*}{\rotatebox{90}{Traffic}} 
    & 96  & \first{0.382} & \first{0.261} & \second{0.395} & \second{0.268} & 0.649 & 0.389 & 0.462 & 0.295 & 0.522 & 0.290 & 0.805 & 0.493 & 0.593 & 0.321 & 0.650 & 0.396 & 0.587 & 0.366 & 0.613 & 0.388  \\
    
    & 192 & \first{0.396} & \first{0.267} & \second{0.417} & \second{0.276} & 0.601 & 0.366 & 0.466 & 0.296 & 0.530 & 0.293 & 0.756 & 0.474 & 0.617 & 0.336 & 0.598 & 0.370 & 0.604 & 0.373 & 0.616 & 0.382  \\
    
    & 336 & \first{0.417} & \first{0.276} & \second{0.433} & \second{0.283} & 0.609 & 0.369 & 0.482 & 0.304 & 0.558 & 0.305 & 0.762 & 0.477 & 0.629 & 0.336 & 0.605 & 0.373 & 0.621 & 0.383 & 0.622 & 0.337  \\
    
    & 720 & \first{0.460} & \first{0.300} & \second{0.467} & \second{0.302} & 0.647 & 0.387 & 0.514 & 0.322 & 0.589 & 0.328 & 0.719 & 0.449 & 0.640 & 0.350 & 0.645 & 0.394 & 0.626 & 0.382 & 0.660 & 0.408  \\
    \cmidrule(lr){2-22}
    
    & Avg & \first{0.414} & \first{0.276} & \second{0.428} & \second{0.282} & 0.626 & 0.378 & 0.481 & 0.304 & 0.550 & 0.304 & 0.760 & 0.473 & 0.620 & 0.336 & 0.625 & 0.383 & 0.610 & 0.376 & 0.628 & 0.379\\
    \midrule
    
    \multirow{4}{*}{\rotatebox{90}{PEMS03}} 
    & 12 & \first{0.065} & \first{0.169} & \second{0.071} & \second{0.174} & 0.126 & 0.236 & 0.099 & 0.216 & 0.090 & 0.203 & 0.178 & 0.305 & 0.085 & 0.192 & 0.122 & 0.243 & 0.126 & 0.251 & 0.272 & 0.385  \\
    
    & 24 & \first{0.087} & \first{0.196} & \second{0.093} & \second{0.201} & 0.246 & 0.334 & 0.142 & 0.259 & 0.121 & 0.240 & 0.257 & 0.371 & 0.118 & 0.223 & 0.201 & 0.317 & 0.149 & 0.275 & 0.334 & 0.440  \\
    
    & 48 & \second{0.133} & \second{0.243} & \first{0.125} & \first{0.236}   & 0.551 & 0.529 & 0.211 & 0.319 & 0.202 & 0.317 & 0.379 & 0.463 & 0.155 & 0.260 & 0.333 & 0.425 & 0.227 & 0.348 & 1.032 & 0.782  \\
    
    & 96 & \second{0.201} & \second{0.305} & \first{0.164} & \first{0.275} & 1.057 & 0.787 & 0.269 & 0.370 & 0.262 & 0.367 & 0.490 & 0.539 & 0.228 & 0.317 & 0.457 & 0.515 & 0.348 & 0.434 & 1.031 & 0.796  \\
    \cmidrule(lr){2-22}
    
    & Avg & \second{0.122} & \second{0.228} & \first{0.113} & \first{0.221} & 0.495 & 0.472 & 0.180 & 0.291 & 0.169 & 0.281 & 0.326 & 0.419 & 0.147 & 0.248 & 0.278 & 0.375 & 0.213 & 0.327 & 0.667 & 0.601\\
    \midrule
    
    \multirow{4}{*}{\rotatebox{90}{PEMS04}} 
    & 12 & \first{0.076} & \first{0.180} & \second{0.078} & \second{0.183} & 0.138 & 0.252 & 0.105 & 0.224 & 0.098 & 0.218 & 0.219 & 0.340 & 0.087 & 0.195 & 0.148 & 0.272 & 0.138 & 0.262 & 0.424 & 0.491\\
    
    & 24 & \first{0.084} & \first{0.193} & \second{0.095} & \second{0.205} & 0.258 & 0.348 & 0.153 & 0.275 & 0.131 & 0.256 & 0.292 & 0.398 & 0.103 & 0.215 & 0.224 & 0.340 & 0.177 & 0.293 & 0.459 & 0.509\\
    
    & 48 & \first{0.115} & \first{0.224} & \second{0.120} & \second{0.233} & 0.572 & 0.544 & 0.229 & 0.339 & 0.205 & 0.326 & 0.409 & 0.478 & 0.136 & 0.250 & 0.355 & 0.437 & 0.270 & 0.368 & 0.646 & 0.610\\
    
    & 96 & \first{0.137} & \first{0.248} & \second{0.150} & \second{0.262} & 1.137 & 0.820 & 0.291 & 0.389 & 0.402 & 0.457 & 0.492 & 0.532 & 0.190 & 0.303 & 0.452 & 0.504 & 0.341 & 0.427 & 0.912 & 0.748\\
    \cmidrule(lr){2-22}
    
    & Avg & \first{0.103} & \first{0.211} & \second{0.111} & \second{0.221} & 0.526 & 0.491 & 0.195 & 0.307 & 0.209 & 0.314 & 0.353 & 0.437 & 0.129 & 0.241 & 0.295 & 0.388 & 0.231 & 0.337 & 0.610 & 0.590\\
    \midrule

    \multirow{4}{*}{\rotatebox{90}{PEMS07}} 
    & 12 & \first{0.063} & \first{0.159} & \second{0.067} & \second{0.165} & 0.118 & 0.235 & 0.095 & 0.207 & 0.094 & 0.200 & 0.173 & 0.304 & 0.082 & 0.181 & 0.115 & 0.242 & 0.109 & 0.225 & 0.199 & 0.336  \\
    
    & 24 & \first{0.081} & \first{0.183} & \second{0.088} & \second{0.190} & 0.242 & 0.341 & 0.150 & 0.262 & 0.139 & 0.247 & 0.271 & 0.383 & 0.101 & 0.204 & 0.210 & 0.329 & 0.125 & 0.244 & 0.323 & 0.420  \\
    
    & 48 & \first{0.093} & \first{0.192} & \second{0.110} & \second{0.215} & 0.562 & 0.541 & 0.253 & 0.340 & 0.311 & 0.369 & 0.446 & 0.495 & 0.134 & 0.238 & 0.398 & 0.458 & 0.165 & 0.288 & 0.390 & 0.470  \\
    
    & 96 & \first{0.117} & \first{0.217} & \second{0.139} & \second{0.245} & 1.096 & 0.795 & 0.346 & 0.404 & 0.396 & 0.442 & 0.628 & 0.577 & 0.181 & 0.279 & 0.594 & 0.553 & 0.262 & 0.376 & 0.554 & 0.578  \\
    \cmidrule(lr){2-22}
    
    & Avg & \first{0.089} & \first{0.188} & \second{0.101} & \second{0.204} & 0.504 & 0.478 & 0.211 & 0.303 & 0.235 & 0.315 & 0.380 & 0.440 & 0.124 & 0.225 & 0.329 & 0.395 & 0.165 & 0.283 & 0.367 & 0.451\\
    \midrule
    
    \multirow{4}{*}{\rotatebox{90}{PEMS08}}
    & 12 & \first{0.076} & \first{0.178} & \second{0.079} & \second{0.182} & 0.133 & 0.247 & 0.168 & 0.232 & 0.165 & 0.214 & 0.227 & 0.343 & 0.112 & 0.212 & 0.154 & 0.276 & 0.173 & 0.273 & 0.436 & 0.485  \\
    
    & 24 & \first{0.104} & \first{0.209} & \second{0.115} & \second{0.219} & 0.249 & 0.343 & 0.224 & 0.281 & 0.215 & 0.260 & 0.318 & 0.409 & 0.141 & 0.238 & 0.248 & 0.353 & 0.210 & 0.301 & 0.467 & 0.502  \\
    
    & 48 & \first{0.167} & \first{0.228} & \second{0.186} & \second{0.235} & 0.569 & 0.544 & 0.321 & 0.354 & 0.315 & 0.355 & 0.497 & 0.510 & 0.198 & 0.283 & 0.440 & 0.470 & 0.320 & 0.394 & 0.966 & 0.733  \\
    
    & 96 & \second{0.245} & \second{0.280} & \first{0.221} & \first{0.267} & 1.166 & 0.814 & 0.408 & 0.417 & 0.377 & 0.397 & 0.721 & 0.592 & 0.320 & 0.351 & 0.674 & 0.565 & 0.442 & 0.465 & 1.385 & 0.915  \\
    \cmidrule(lr){2-22}
    
    & Avg & \first{0.148} & \first{0.224} & \second{0.150} & \second{0.226} & 0.529 & 0.487 & 0.280 & 0.321 & 0.268 & 0.307 & 0.441 & 0.464 & 0.193 & 0.271 & 0.379 & 0.416 & 0.286 & 0.358 & 0.814 & 0.659\\
    \bottomrule
  \end{tabular}
  \end{small}
  \end{threeparttable}
  }
\end{table*}

Our models are fairly compared with 9 representative and state-of-the-art (SOTA) forecasting models, including (1) Transformer-based methods: iTransformer \citep{liu2023itransformer}, PatchTST \citep{nie2022patchtst}, Crossformer \citep{zhang2022crossformer}, FEDformer \citep{zhou2022fedformer}, Autoformer \citep{wu2021autoformer}; (2) Linear-based methods: RLinear \citep{li2023rlinear}, TiDE \citep{das2023tide}, DLinear \citep{zeng2023dlinear}; and (3) Temporal Convolutional Network-based methods: TimesNet \citep{wu2022timesnet}. 
The brief introductions of these models are as follows: 
\begin{itemize}
    \item iTransformer reverses the order of information processing, which first analyzes the time series information of each individual variate and then fuses the information of all variates. This unique approach has positioned iTransformer as the current SOTA model in TSF. 
    \item PatchTST segments time series into subseries patches as input tokens and uses channel-independent shared embeddings and weights for efficient representation learning.
    \item Crossformer introduces a cross-attention mechanism that allows the model to interact with information between different time steps to help the model capture long-term dependencies in time series. 
    \item FEDformer is a frequency-enhanced Transformer that takes advantage of the fact that most time series tend to have a sparse representation in well-known basis such as Fourier transform to improve performance. 
    \item Autoformer takes a decomposition architecture that incorporates an auto-correlation mechanism and updates traditional sequence decomposition into the basic inner blocks of the depth model. 
    \item RLinear is the SOTA linear model, which employs reversible normalization and channel independence into pure linear structure.
    \item TiDE is a Multi-layer Perceptron (MLP) based encoder-decoder model.
    \item DLinear is the first linear model in TSF and a simple one-layer linear model with decomposition architecture. 
    \item TimesNet uses TimesBlock as a task-general backbone, transforms 1D time series into 2D tensors, and captures intraperiod and interperiod variations using 2D kernels. 
\end{itemize}

\begin{table*}[htbp]
  \caption{Full results of S-Mamba and baselines on ETT datasets.
  The lookback length $L$ is set to 96 and the forecast length $T$ is set to 96, 192, 336, 720. }
  \label{tab:results_ETT}
  \centering
  \renewcommand{\arraystretch}{0.9}
  \resizebox{\textwidth}{!}{
  \begin{threeparttable}
  \begin{small}
  \setlength{\tabcolsep}{2.6pt}
  % \vspace{-3mm}
  \begin{tabular}{c|c|cc|cc|cc|cc|cc|cc|cc|cc|cc|cc}
    \toprule
    \multicolumn{2}{c|}{Models} & \multicolumn{2}{c|}{\textbf{S-Mamba}} & \multicolumn{2}{c}{iTransformer} & \multicolumn{2}{c}{RLinear} & \multicolumn{2}{c}{PatchTST} & \multicolumn{2}{c}{Crossformer} & \multicolumn{2}{c}{TiDE} & \multicolumn{2}{c}{TimesNet} & \multicolumn{2}{c}{DLinear} & \multicolumn{2}{c}{FEDformer} &  \multicolumn{2}{c}{Autoformer} \\
    \cmidrule(lr){1-2}\cmidrule(lr){3-4}\cmidrule(lr){5-6}\cmidrule(lr){7-8} \cmidrule(lr){9-10}\cmidrule(lr){11-12}\cmidrule(lr){13-14}\cmidrule(lr){15-16}\cmidrule(lr){17-18}\cmidrule(lr){19-20}\cmidrule(lr){21-22}  
    \multicolumn{2}{c|}{Metric} & MSE & MAE & MSE & MAE & MSE & MAE & MSE & MAE & MSE & MAE & MSE & MAE & MSE & MAE & MSE & MAE & MSE & MAE & MSE & MAE  \\
    \toprule
    % & 720 &  &  &  &  &  &  &  &  &  &  &  &  &  &  \\
    \multirow{4}{*}{\rotatebox{90}{ETTm1}} 
    &  96 & \second{0.333} & \second{0.368} & 0.334 & \second{0.368} & 0.355 & 0.376 & \first{0.329} & \first{0.367} & 0.404 & 0.426 & 0.364 & 0.387 & 0.338 & 0.375 & 0.345 & 0.372 & 0.379 & 0.419 & 0.505 & 0.475\\
    
    & 192 & 0.376 & 0.390 & 0.377 & 0.391 & 0.391 & 0.392 & \first{0.367} & \first{0.385} & 0.450 & 0.451 & 0.398 & 0.404 & \second{0.374} & \second{0.387} & 0.380 & 0.389 & 0.426 & 0.441 & 0.553 & 0.496\\
    
    & 336 & 0.408 & 0.413 & 0.426 & 0.420 & 0.424 & 0.415 & \first{0.399} & \first{0.410} & 0.532 & 0.515 & 0.428 & 0.425 & \second{0.410} & \second{0.411} & 0.413 & 0.413 & 0.445 & 0.459 & 0.621 & 0.537\\
    
    & 720 & 0.475 & \second{0.448} & 0.491 & 0.459 & 0.487 & 0.450 & \first{0.454} & \first{0.439} & 0.666 & 0.589 & 0.487 & 0.461 & 0.478 & 0.450 & \second{0.474} & 0.453 & 0.543 & 0.490 & 0.671 & 0.561\\
    \cmidrule(lr){2-22}

    & Avg & \second{0.398} & \second{0.405} & 0.407 & 0.410 & 0.414 & 0.407 & \first{0.387} & \first{0.400} & 0.513 & 0.496 & 0.419 & 0.419 & 0.400 & 0.406 & 0.403 & 0.407 & 0.448 & 0.452 & 0.588 & 0.517\\
    \midrule
    
    \multirow{4}{*}{\rotatebox{90}{ETTm2}} 
    & 96  & \second{0.179} & \second{0.263} & 0.180 & 0.264 & 0.182 & 0.265 & \first{0.175} & \first{0.259} & 0.287 & 0.366 & 0.207 & 0.305 & 0.187 & 0.267 & 0.193 & 0.292 & 0.203 & 0.287 & 0.255 & 0.339  \\
    
    & 192 & 0.250 & 0.309 & 0.250 & 0.309 & \second{0.246} & \second{0.304} & \first{0.241} & \first{0.302} & 0.414 & 0.492 & 0.290 & 0.364 & 0.249 & 0.309 & 0.284 & 0.362 & 0.269 & 0.328 & 0.281 & 0.340  \\
    
    & 336 & 0.312 & 0.349 & 0.311 & 0.348 & \second{0.307} & \first{0.342} & \first{0.305} & \second{0.343} & 0.597 & 0.542 & 0.377 & 0.422 & 0.321 & 0.351 & 0.369 & 0.427 & 0.325 & 0.366 & 0.339 & 0.372  \\
    
    & 720 & 0.411 & 0.406 & 0.412 & 0.407 & \second{0.407} & \first{0.398} & \first{0.402} & \second{0.400} & 1.730 & 1.042 & 0.558 & 0.524 & 0.408 & 0.403 & 0.554 & 0.522 & 0.421 & 0.415 & 0.433 & 0.432  \\
    \cmidrule(lr){2-22}

    & Avg & 0.288 & 0.332 & 0.288 & 0.332 & \second{0.286} & \second{0.327} & \first{0.281} & \first{0.326} & 0.757 & 0.610 & 0.358 & 0.404 & 0.291 & 0.333 & 0.350 & 0.401 & 0.305 & 0.349 & 0.327 & 0.371\\
    
    \midrule
    \multirow{5}{*}{\rotatebox{90}{ETTh1}} 
    &  96 & 0.386 & 0.405 & 0.386 & 0.405 & 0.386 & \first{0.395} & 0.414 & 0.419 & 0.423 & 0.448 & 0.479 & 0.464 & \second{0.384} & 0.402 & 0.386 & \second{0.400} & \first{0.376} & 0.419 & 0.449 & 0.459  \\
    
    & 192 & 0.443 & 0.437 & 0.441 & 0.436 & 0.437 & \first{0.424} & 0.460 & 0.445 & 0.471 & 0.474 & 0.525 & 0.492 & \second{0.436} & \second{0.429} & 0.437 & 0.432 & \first{0.420} & 0.448 & 0.500 & 0.482  \\
    
    & 336 & 0.489 & 0.468 & 0.487 & \second{0.458} & \second{0.479} & \first{0.446} & 0.501 & 0.466 & 0.570 & 0.546 & 0.565 & 0.515 & 0.491 & 0.469 & 0.481 & 0.459 & \first{0.459} & 0.465 & 0.521 & 0.496  \\
    
    & 720 & 0.502 & 0.489 & 0.503 & 0.491 & \first{0.481} & \first{0.470} & \second{0.500} & \second{0.488} & 0.653 & 0.621 & 0.594 & 0.558 & 0.521 & 0.500 & 0.519 & 0.516 & 0.506 & 0.507 & 0.514 & 0.512  \\
    \cmidrule(lr){2-22}

    & Avg & 0.455 & 0.450 & 0.454 & \second{0.447} & \second{0.446} & \first{0.434} & 0.469 & 0.454 & 0.529 & 0.522 & 0.541 & 0.507 & 0.458 & 0.450 & 0.456 & 0.452 & \first{0.440} & 0.460 & 0.496 & 0.487\\
    \midrule
    
    \multirow{5}{*}{\rotatebox{90}{ETTh2}} 
    & 96  & \second{0.296} & \second{0.348} & 0.297 & 0.349 & \first{0.288} & \first{0.338} & 0.302 & 0.348 & 0.745 & 0.584 & 0.400 & 0.440 & 0.340 & 0.374 & 0.333 & 0.387 & 0.358 & 0.397 & 0.346 & 0.388  \\
    
    & 192 & \second{0.376} & \second{0.396} & 0.380 & 0.400 & \first{0.374} & \first{0.390} & 0.388 & 0.400 & 0.877 & 0.656 & 0.528 & 0.509 & 0.402 & 0.414 & 0.477 & 0.476 & 0.429 & 0.439 & 0.456 & 0.452  \\
    
    & 336 & \second{0.424} & \second{0.431} & 0.428 & 0.432 & \first{0.415} & \first{0.426} & 0.426 & 0.433 & 1.043 & 0.731 & 0.643 & 0.571 & 0.452 & 0.452 & 0.594 & 0.541 & 0.496 & 0.487 & 0.482 & 0.486  \\
    
    & 720 & \second{0.426} & \second{0.444} & 0.427 & 0.445 & \first{0.420} & \first{0.440} & 0.431 & 0.446 & 1.104 & 0.763 & 0.874 & 0.679 & 0.462 & 0.468 & 0.831 & 0.657 & 0.463 & 0.474 & 0.515 & 0.511  \\
    \cmidrule(lr){2-22}

    & Avg & \second{0.381} & \second{0.405} & 0.383 & 0.407 & \first{0.374} & \first{0.398} & 0.387 & 0.407 & 0.942 & 0.684 & 0.611 & 0.550 & 0.414 & 0.427 & 0.559 & 0.515 & 0.437 & 0.449 & 0.450 & 0.459\\
    \bottomrule
  \end{tabular}
  
  \end{small}
  \end{threeparttable}
  }
\end{table*}

\begin{table*}[htbp]
  \caption{Full results of S-Mamba and baselines on Electricity, Exchange, Weather and Solar-Energy datasets.
  The lookback length $L$ is set to 96 and the forecast length $T$ is set to 96, 192, 336, 720.}
  \label{tab:results_other}
  \renewcommand{\arraystretch}{0.9}
  \centering
  \resizebox{\textwidth}{!}{
  \begin{threeparttable}
  \begin{small}
  \setlength{\tabcolsep}{2.6pt}
  % \vspace{-3mm}
  \begin{tabular}{c|c|cc|cc|cc|cc|cc|cc|cc|cc|cc|cc}
    \toprule
    \multicolumn{2}{c|}{Models} & \multicolumn{2}{c|}{\textbf{S-Mamba}} & \multicolumn{2}{c}{iTransformer} & \multicolumn{2}{c}{RLinear} & \multicolumn{2}{c}{PatchTST} & \multicolumn{2}{c}{Crossformer} & \multicolumn{2}{c}{TiDE} & \multicolumn{2}{c}{TimesNet} & \multicolumn{2}{c}{DLinear} & \multicolumn{2}{c}{FEDformer} &  \multicolumn{2}{c}{Autoformer} \\
    \cmidrule(lr){1-2}\cmidrule(lr){3-4}\cmidrule(lr){5-6}\cmidrule(lr){7-8} \cmidrule(lr){9-10}\cmidrule(lr){11-12}\cmidrule(lr){13-14}\cmidrule(lr){15-16}\cmidrule(lr){17-18}\cmidrule(lr){19-20}\cmidrule(lr){21-22}  
    \multicolumn{2}{c|}{Metric} & MSE & MAE & MSE & MAE & MSE & MAE & MSE & MAE & MSE & MAE & MSE & MAE & MSE & MAE & MSE & MAE & MSE & MAE & MSE & MAE  \\
    \toprule
    \multirow{5}{*}{\rotatebox{90}{Electricity}} 
    & 96  & \first{0.139} & \first{0.235} & \second{0.148} & \second{0.240} & 0.201 & 0.281 & 0.181 & 0.270 & 0.219 & 0.314 & 0.237 & 0.329 & 0.168 & 0.272 & 0.197 & 0.282 & 0.193 & 0.308 & 0.201 & 0.317\\
    
    & 192 & \first{0.159} & \second{0.255} & \second{0.162} & \first{0.253} & 0.201 & 0.283 & 0.188 & 0.274 & 0.231 & 0.322 & 0.236 & 0.330 & 0.184 & 0.289 & 0.196 & 0.285 & 0.201 & 0.315 & 0.222 & 0.334\\
    
    & 336 & \first{0.176} & \second{0.272} & \second{0.178} & \first{0.269} & 0.215 & 0.298 & 0.204 & 0.293 & 0.246 & 0.337 & 0.249 & 0.344 & 0.198 & 0.300 & 0.209 & 0.301 & 0.214 & 0.329 & 0.231 & 0.338\\
    
    & 720 & \first{0.204} & \first{0.298} & 0.225 & \second{0.317} & 0.257 & 0.331 & 0.246 & 0.324 & 0.280 & 0.363 & 0.284 & 0.373 & \second{0.220} & 0.320 & 0.245 & 0.333 & 0.246 & 0.355 & 0.254 & 0.361\\
    \cmidrule(lr){2-22}
    
    & Avg & \first{0.170} & \first{0.265} & \second{0.178} & \second{0.270} & 0.219 & 0.298 & 0.205 & 0.290 & 0.244 & 0.334 & 0.251 & 0.344 & 0.192 & 0.295 & 0.212 & 0.300 & 0.214 & 0.327 & 0.227 & 0.338\\
    \midrule
    
    \multirow{5}{*}{\rotatebox{90}{Exchange}} 
    & 96  & \first{0.086} & 0.207 & \first{0.086} & \second{0.206} & 0.093 & 0.217 & \second{0.088} & \first{0.205} & 0.256 & 0.367 & 0.094 & 0.218 & 0.107 & 0.234 & \second{0.088} & 0.218 & 0.148 & 0.278 & 0.197 & 0.323\\
    
    & 192 & 0.182 & 0.304 & \second{0.177} & \first{0.299} & 0.184 & 0.307 & \first{0.176} & \first{0.299} & 0.470 & 0.509 & 0.184 & 0.307 & 0.226 & 0.344 & \first{0.176} & 0.315 & 0.271 & 0.315 & 0.300 & 0.369\\
    
    & 336 & 0.332 & 0.418 & 0.331 & \second{0.417} & 0.351 & 0.432 & \first{0.301} & \first{0.397} & 1.268 & 0.883 & 0.349 & 0.431 & 0.367 & 0.448 & \second{0.313} & 0.427 & 0.460 & 0.427 & 0.509 & 0.524\\
    
    & 720 & 0.867 & 0.703 & \second{0.847} & \first{0.691} & 0.886 & 0.714 & 0.901 & 0.714 & 1.767 & 1.068 & 0.852 & 0.698 & 0.964 & 0.746 & \first{0.839} & \second{0.695} & 1.195 & \second{0.695} & 1.447 & 0.941\\
    
    & Avg & 0.367 & 0.408 & \second{0.360} & \first{0.403} & 0.378 & 0.417 & 0.367 & \second{0.404} & 0.940 & 0.707 & 0.370 & 0.413 & 0.416 & 0.443 & \first{0.354} & 0.414 & 0.519 & 0.429 & 0.613 & 0.539\\
    \midrule
    
    \multirow{5}{*}{\rotatebox{90}{Weather}} 
    & 96  & \second{0.165} & \first{0.210} & 0.174 & \second{0.214} & 0.192 & 0.232 & 0.177 & 0.218 & \first{0.158} & 0.230 & 0.202 & 0.261 & 0.172 & 0.220 & 0.196 & 0.255 & 0.217 & 0.296 & 0.266 & 0.336\\
    
    & 192 & \second{0.214} & \first{0.252} & 0.221 & \second{0.254} & 0.240 & 0.271 & 0.225 & 0.259 & \first{0.206} & 0.277 & 0.242 & 0.298 & 0.219 & 0.261 & 0.237 & 0.296 & 0.276 & 0.336 & 0.307 & 0.367\\
    
    & 336 & \second{0.274} & \second{0.297} & 0.278 & \first{0.296} & 0.292 & 0.307 & 0.278 & \second{0.297} & \first{0.272} & 0.335 & 0.287 & 0.335 & 0.280 & 0.306 & 0.283 & 0.335 & 0.339 & 0.380 & 0.359 & 0.395\\
    
    & 720 & \second{0.350} & \first{0.345} & 0.358 & \second{0.347} & 0.364 & 0.353 & 0.354 & 0.348 & 0.398 & 0.418 & 0.351 & 0.386 & 0.365 & 0.359 & \first{0.345} & 0.381 & 0.403 & 0.428 & 0.419 & 0.428\\
    \cmidrule(lr){2-22}
    
    & Avg & \first{0.251} & \first{0.276} & \second{0.258} & \second{0.278} & 0.272 & 0.291 & 0.259 & 0.281 & 0.259 & 0.315 & 0.271 & 0.320 & 0.259 & 0.287 & 0.265 & 0.317 & 0.309 & 0.360 & 0.338 & 0.382\\
    \midrule
    
    \multirow{5}{*}{\rotatebox{90}{Solar-Energy}} 
    & 96  & \second{0.205} & \second{0.244} & \first{0.203} & \first{0.237} & 0.322 & 0.339 & 0.234 & 0.286 & 0.310 & 0.331 & 0.312 & 0.399 & 0.250 & 0.292 & 0.290 & 0.378 & 0.242 & 0.342 & 0.884 & 0.711  \\
    
    & 192 & \second{0.237} & \second{0.270} & \first{0.233} & \first{0.261} & 0.359 & 0.356 & 0.267 & 0.310 & 0.734 & 0.725 & 0.339 & 0.416 & 0.296 & 0.318 & 0.320 & 0.398 & 0.285 & 0.380 & 0.834 & 0.692  \\
    
    & 336 & \second{0.258} & \second{0.288} & \first{0.248} & \first{0.273} & 0.397 & 0.369 & 0.290 & 0.315 & 0.750 & 0.735 & 0.368 & 0.430 & 0.319 & 0.330 & 0.353 & 0.415 & 0.282 & 0.376 & 0.941 & 0.723  \\
    
    & 720 & \second{0.260} & \second{0.288} & \first{0.249} & \first{0.275} & 0.397 & 0.356 & 0.289 & 0.317 & 0.769 & 0.765 & 0.370 & 0.425 & 0.338 & 0.337 & 0.356 & 0.413 & 0.357 & 0.427 & 0.882 & 0.717  \\
    \cmidrule(lr){2-22}
    
    & Avg & \second{0.240} & \second{0.273} & \first{0.233} & \first{0.262} & 0.369 & 0.356 & 0.270 & 0.307 & 0.641 & 0.639 & 0.347 & 0.417 & 0.301 & 0.319 & 0.330 & 0.401 & 0.291 & 0.381 & 0.885 & 0.711\\
    \bottomrule
  \end{tabular}
  \end{small}
  \end{threeparttable}
  }
\end{table*}

% The results of baselines are reported by \citet{liu2023itransformer}.
\subsection{Overall Performance}
Tab. \ref{tab:results_traffic}, Tab. \ref{tab:results_ETT}, and Tab. \ref{tab:results_other} present a comparative analysis of the overall performance of our models and other baseline models across all datasets. 
The best results are highlighted in bold red font, while the second best results are presented in underlined purple font. 
From the data presented in these tables, we summarize three observations and attach the analysis: 
(1) S-Mamba attain commendable outcomes on the traffic-related, Electricity, and Solar-Energy datasets. 
These datasets are distinguished by their numerous variates, most of which are periodic. 
It is worth noting that period variates are more likely to contain learnable VC. 
Mamba VC Fusion Layer benefits from this characteristic and improves S-Mamba performance. 
(2) In the context of the ETT, and Exchange datasets, S-Mamba does not demonstrate a pronounced superiority in performance; indeed, it exhibits a suboptimal outcome. 
This can be attributed to the fact that these datasets are characterized by a few number of variates, predominantly of an aperiodic nature. 
Consequently, there are weak VCs between these variates, and the employment of Mamba VC Encoding layer by S-Mamba can't bring useful information and even may inadvertently introduce noise into the predictive model, thus impeding its predictive accuracy. 
(3) The Weather dataset is special in that it has fewer variates and most variates are aperiodic, but S-Mamba still achieves the best performance on it. 
We think that this phenomenon arises from the tendency of variates in the Weather dataset to exhibit simultaneous trends of either falling or rising despite the absence of periodic patterns So the Mamba VC Encoding Layer of S-Mamba can still benefit from these data. 
Moreover, the Weather dataset exhibits large sections of rising or falling trends. The Feed-Forward Network (FFN) layer accurately records these relationships, which is also beneficial for S-Mamba’s comprehension.

\begin{figure*}
  \centering
  \subcaptionbox{S-Mamba}{
  \includegraphics[width=0.2\linewidth]{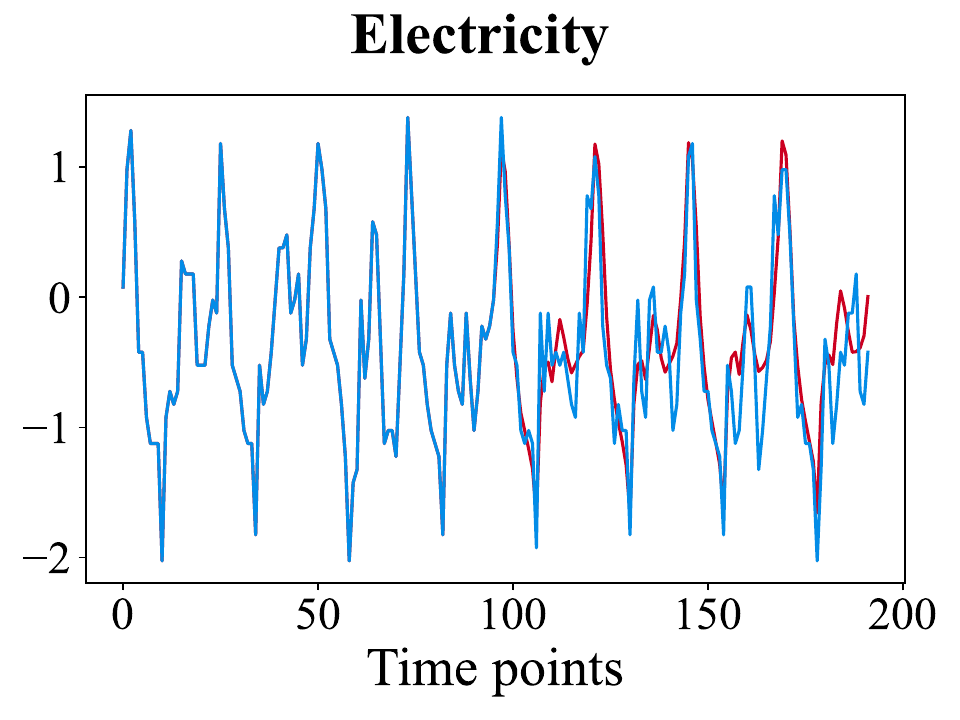}
  \includegraphics[width=0.2\linewidth]{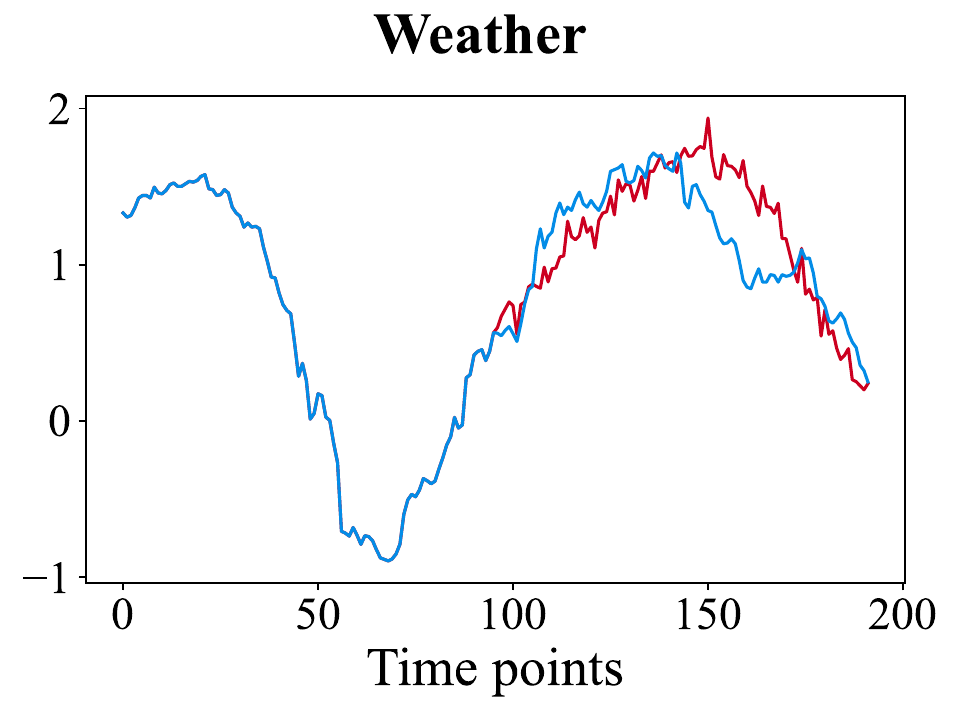}
  \includegraphics[width=0.2\linewidth]{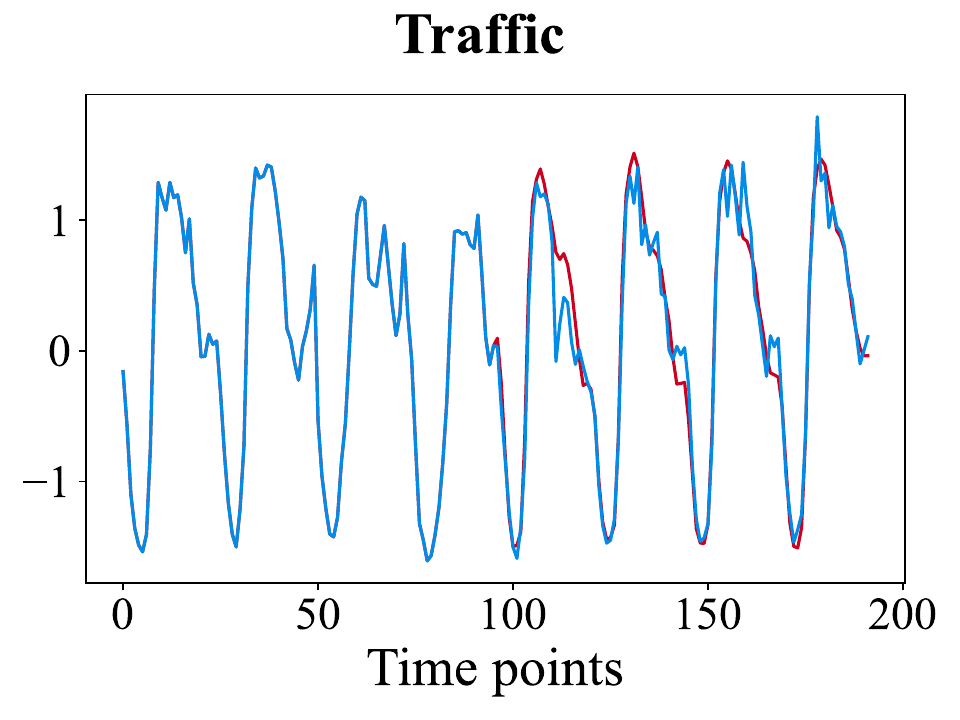}
  \includegraphics[width=0.2\linewidth]{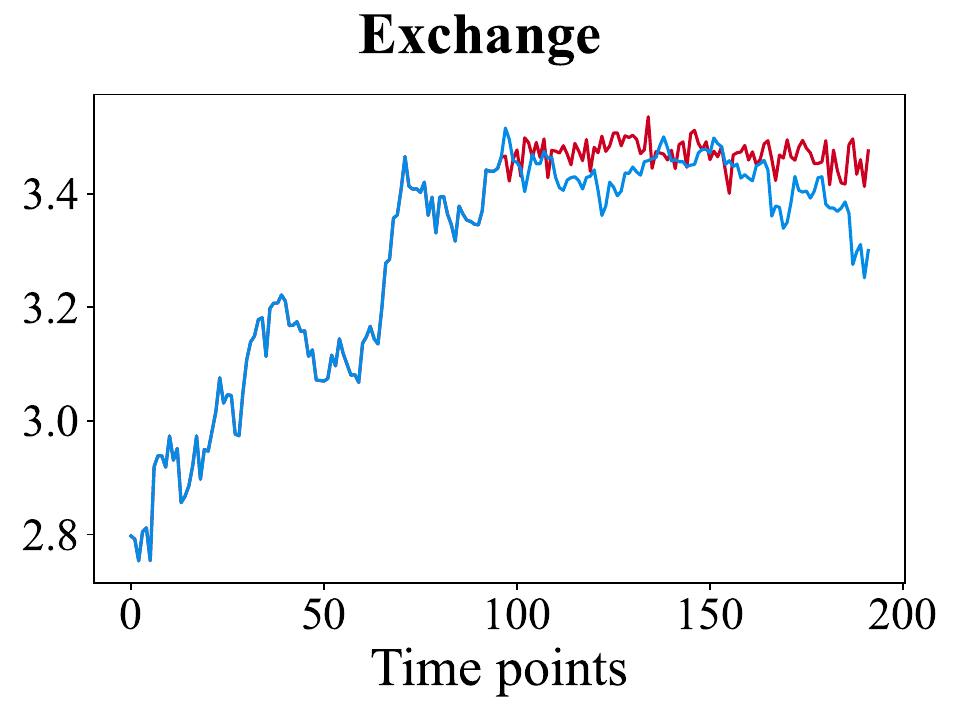}
  \includegraphics[width=0.2\linewidth]{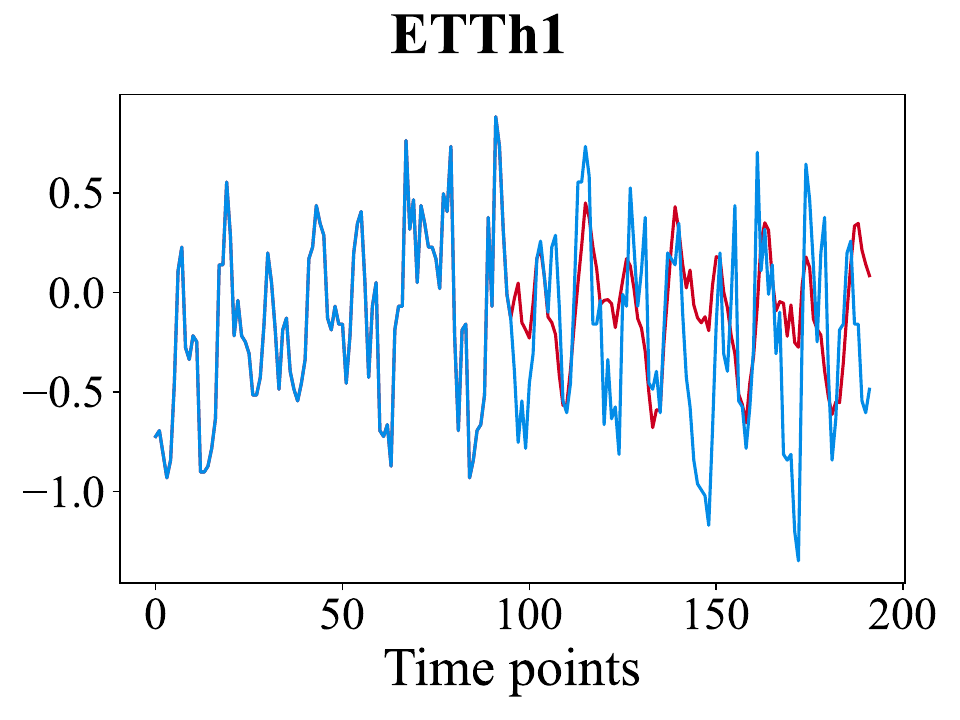}
  }
  
  \subcaptionbox{iTransformer}{
  \includegraphics[width=0.2\linewidth]{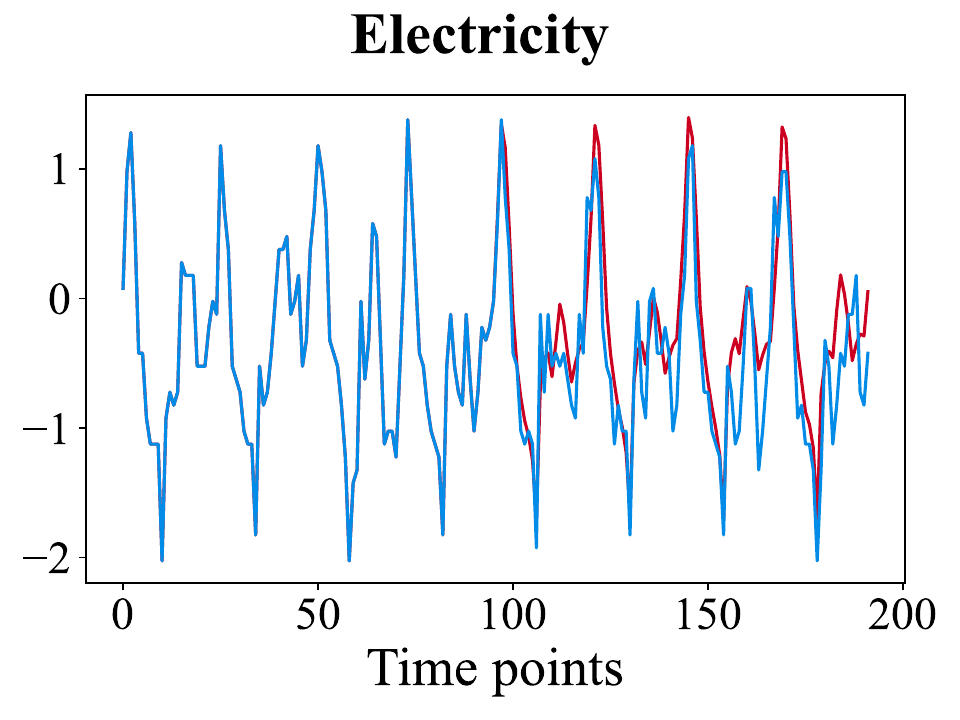}
  \includegraphics[width=0.2\linewidth]{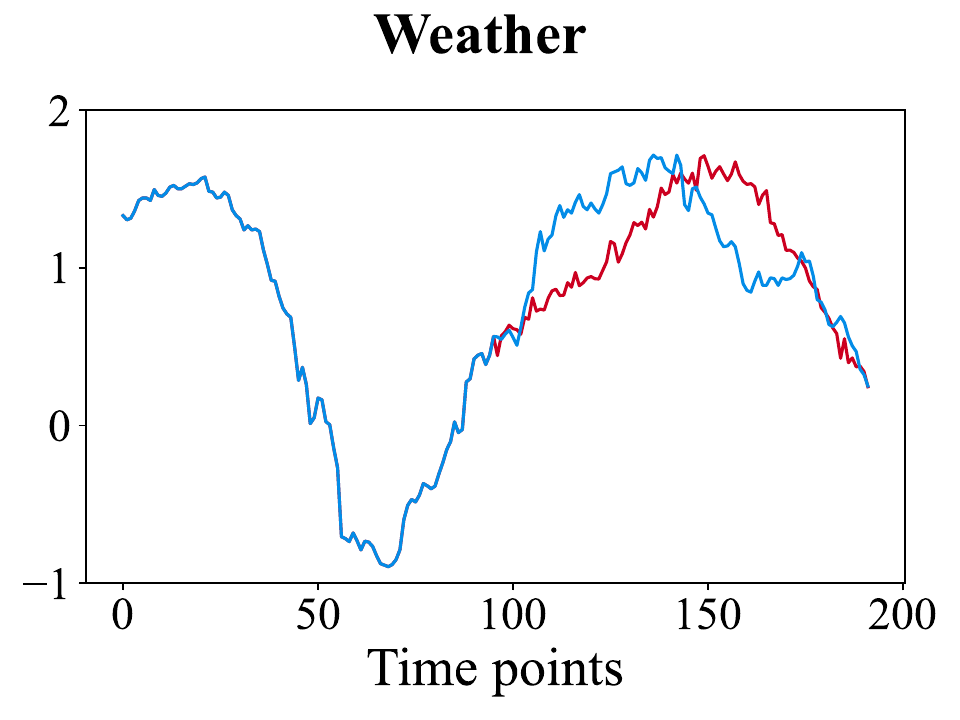}
  \includegraphics[width=0.2\linewidth]{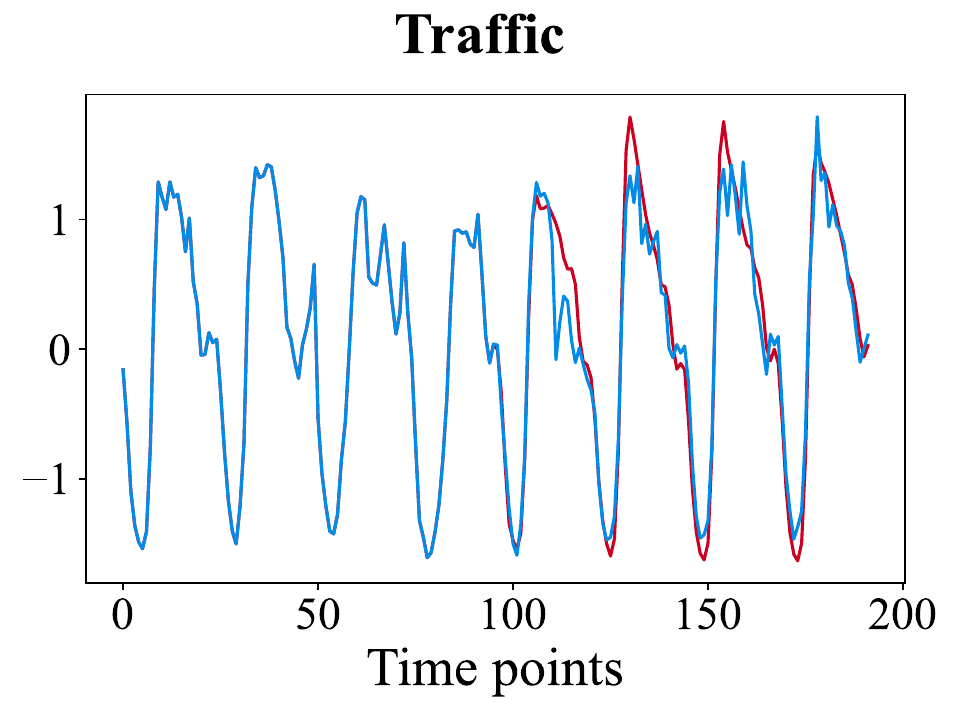}
  \includegraphics[width=0.2\linewidth]{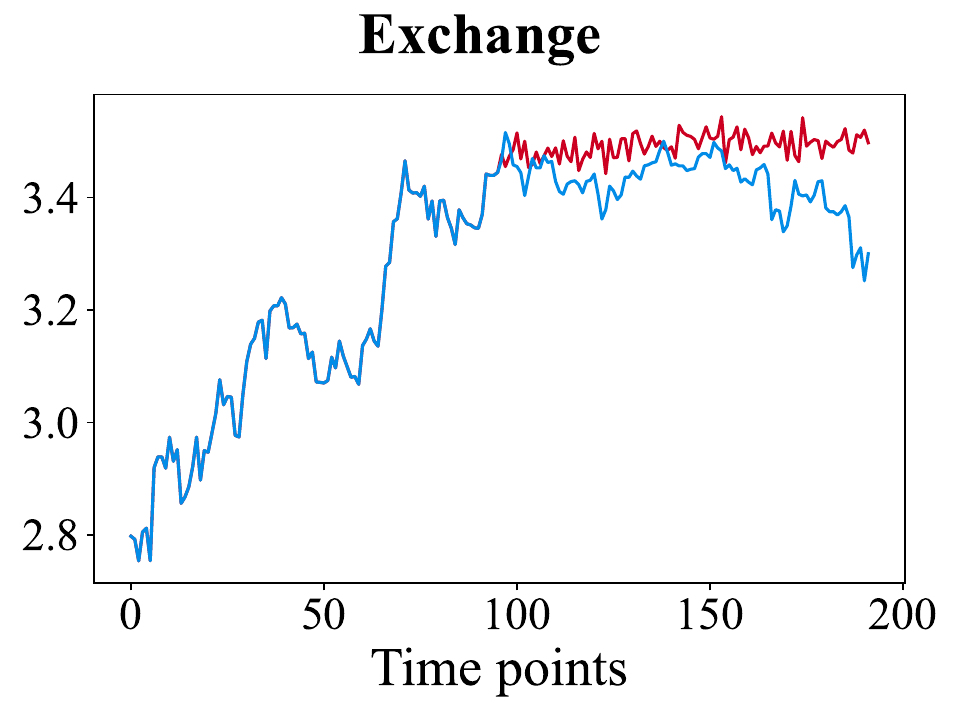}
  \includegraphics[width=0.2\linewidth]{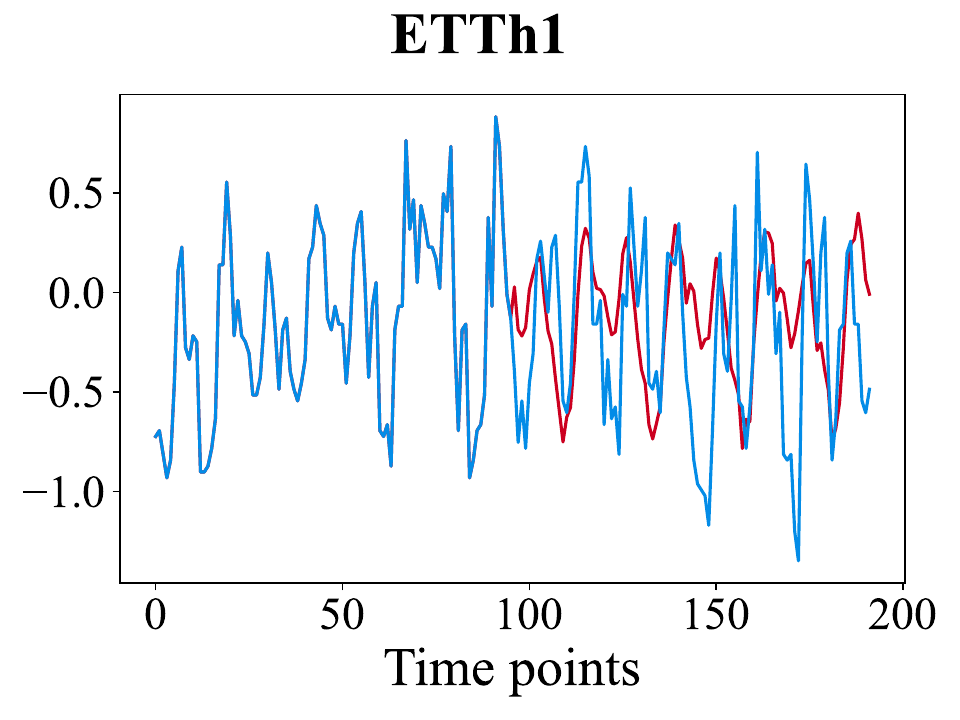}
  }
  \caption{Comparison of forecasts between S-Mamba and iTransformer on five datasets when the input length is 96 and the forecast length is 96. 
  The blue line represents the ground truth and the red line represents the forecast.}
  \label{fig:line}
\end{figure*}

Furthermore, to provide a more intuitive assessment of S-Mamba's forecast capabilities, we visually compare the predictions of S-Mamba and the leading baseline, iTransformer, on four datasets: Electricity, Weather, Traffic, Exchange, and ETTh1, through graphical representation. 
Specifically, we randomly select a variate and then input its lookback sequence, where the true subsequent sequence is depicted as a blue line and the model's forecast is represented by a red line in Fig. \ref{fig:line}. 
It is evident that on the Electricity, Weather, and Traffic datasets, S-Mamba's predictions closely approximate the actual values, with nearly perfect alignment observed on the Electricity and Traffic datasets and are better than iTransformer.  
On the Exchange and ETTh1, the two models exhibit similar performance because the two datasets contain few variates, so there is no evident gap between using bidirectional Mamba or using Transformer for information fusion between variates.

\begin{figure*}[!htbp]
 \centering
 \includegraphics[width=0.9\textwidth]{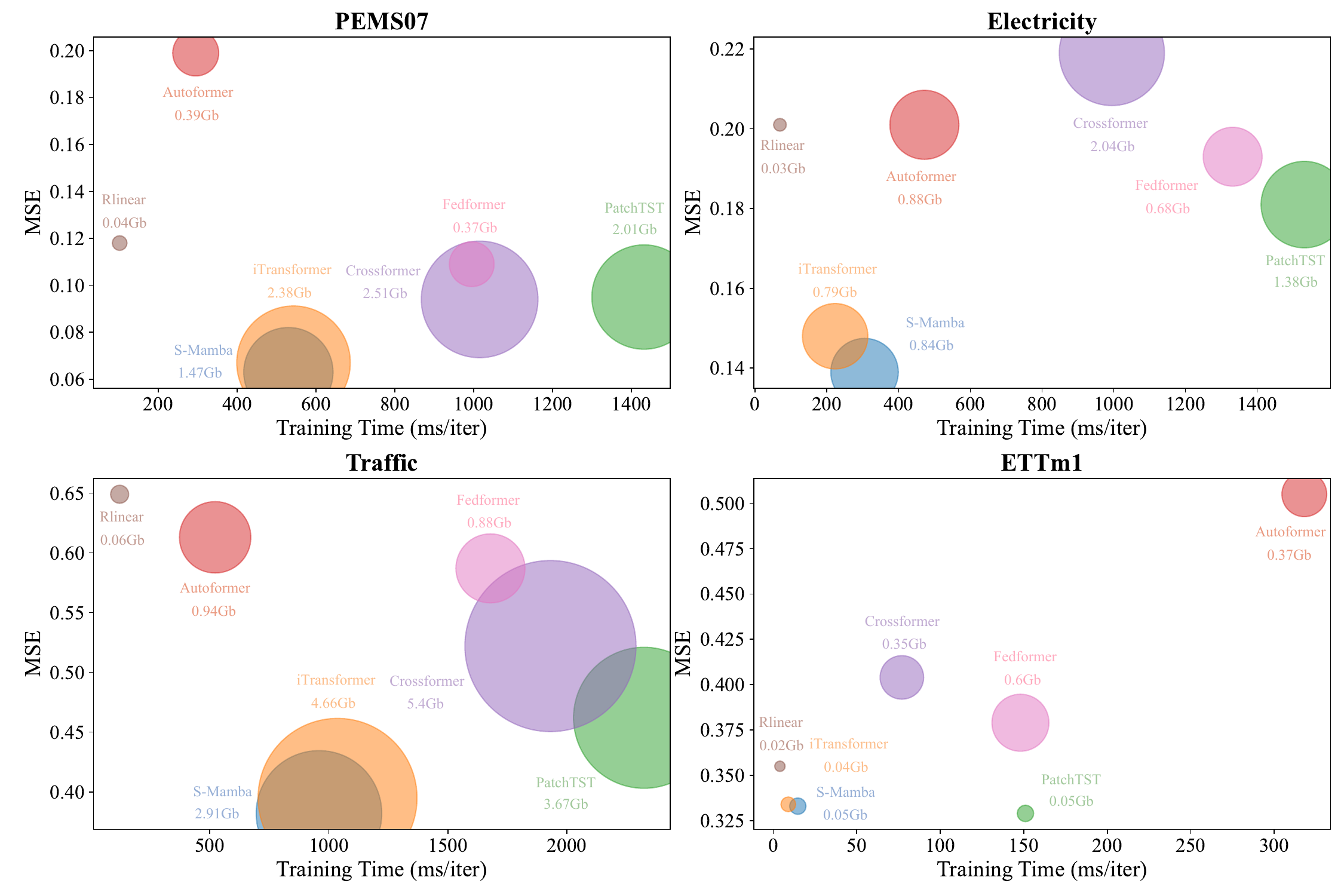}
 \caption{Comparison of S-Mamba and six baselines on MSE, Training Time, and GPU Memory. The lookback length $L=96$, and the forecast length $T=12$ for PEMS07 and $T=96$ for other datasets}
 \label{fig:Efficiency}
 \vspace{-5mm}
\end{figure*}

\subsection{Model Efficiency}
To evaluate the computational efficiency of the models, we compare the memory usage and computing time of S-Mamba with several baselines on PEMS07, Electricity, Traffic, and ETTm1. 
Independent runs are conducted on a single NVIDIA RTX3090 GPU with the batch size set to $16$ and meticulously document the results in Fig. \ref{fig:Efficiency}. 
In our analysis, bubble charts are used to depict the measurement outcomes, wherein the vertical axis denotes the Mean Squared Error (MSE), the horizontal axis quantifies the training duration, and the bubble magnitude correlates with the allocated GPU memory. 
The visualization reveals that the S-Mamba algorithm attains the most favorable MSE metric across the PEMS07, Electricity, and Traffic datasets. 
When benchmarked against Transformer-based models, S-Mamba typically necessitates short training time and low allocated GPU memory. 
The RLinear model does utilize minimal GPU memory and curtails training time, it does not confer a competitive edge in terms of forecast precision. 
Overall, S-Mamba manifests exemplary predictive accuracy with a low computational resource footprint. 

\begin{table*}[!h]
  \caption{Ablation study on Electricity, Traffic, Weather, Solar-Energy, and ETTh2. The lookback length $L=96$, while the forecast length $T \in \left\{96, 192, 336, 720\right\}$. }
  \label{tab:full_ab}
  \centering
  \resizebox{0.85\linewidth}{!}{
  \renewcommand{\arraystretch}{0.9}
  \begin{threeparttable}
  \begin{small}
  \renewcommand{\multirowsetup}{\centering}
  \setlength{\tabcolsep}{3.0pt}
  \begin{tabular}{c|c|c|c|cc|cc|cc|cc|cc}
    \toprule
    \multirow{2}{*}{Design} &\multirow{2}{*}{VC Encoding} & \multirow{2}{*}{TD Encoding} & Forecast & \multicolumn{2}{c|}{Electricity} & \multicolumn{2}{c|}{Traffic} & \multicolumn{2}{c|}{Weather} & \multicolumn{2}{c}{Solar-Energy}& \multicolumn{2}{c}{ETTh2}\\
    \cmidrule(lr){5-6} \cmidrule(lr){7-8}  \cmidrule(lr){9-10} \cmidrule(lr){11-12} \cmidrule(lr){13-14} 
    & & &Lengths & MSE & MAE & MSE & MAE  & MSE & MAE & MSE & MAE & MSE & MAE\\
    \midrule
    \multirow{4}{*}{\textbf{S-Mamba}} & \multirow{4}{*}{\textbf{bi-Mamba}} & \multirow{4}{*}{\textbf{FFN}}
    &      96 & \first{0.139} & \first{0.235} & \second{0.382} & \second{0.261} & 0.165 & 0.210 & \second{0.205} & \second{0.244} & \first{0.296} & \first{0.348}\\
    & & & 192 & \first{0.159} & \first{0.255} & \first{0.396} & \first{0.267} & 0.214 & 0.252 & \second{0.237} & \second{0.270} & \first{0.376} & \first{0.396}\\
    & & & 336 & \first{0.176} & \second{0.272} & \first{0.417} & \first{0.276} & 0.274 & 0.297 & \second{0.258} & \second{0.288} & \first{0.424} & \first{0.431}\\
    & & & 720 & \first{0.204} & \first{0.298} & \first{0.460} & \second{0.300} & 0.350 & 0.345 & \second{0.260} & \second{0.288} & \first{0.426} & \first{0.444}\\
    \midrule
    \multirow{16}{*}{Replace} 
    & \multirow{4}{*}{bi-Mamba} & \multirow{4}{*}{uni-Mamba}
    &      96 & 0.155 & 0.260 & 0.488 & 0.329 & \first{0.161} & \first{0.204} & 0.213 & 0.255 & \second{0.297} & \second{0.349}\\
    & & & 192 & 0.173 & 0.271 & 0.511 & 0.341 & \first{0.208} & \first{0.249} & 0.247 & 0.280 & 0.378 & 0.399\\
    & & & 336 & 0.188 & 0.281 & 0.531 & 0.347 & \first{0.265} & \first{0.280} & 0.267 & 0.298 & 0.428 & 0.437\\
    & & & 720 & 0.210 & 0.308 & 0.621 & 0.352 & \first{0.343} & \first{0.339} & 0.272 & 0.295 & 0.436 & 0.451\\
    \cmidrule(lr){2-14}
    & \multirow{4}{*}{bi-Mamba} & \multirow{4}{*}{bi-Mamba}
    &      96 & 0.154 & 0.259 & 0.512 & 0.348 & \second{0.162} & \second{0.205} & 0.221 & 0.261 & \second{0.297} & \second{0.349}\\
    & & & 192 & 0.175 & 0.273 & 0.505 & 0.344 & \second{0.210} & \second{0.250} & 0.271 & 0.291 & \second{0.377} & \second{0.398}\\
    & & & 336 & 0.184 & 0.276 & 0.527 & 0.369 & \second{0.266} & \second{0.288} & 0.271 & 0.291 & 0.428 & 0.437\\
    & & & 720 & 0.216 & 0.315 & 0.661 & 0.423 & \second{0.344} & \first{0.339} & 0.278 & 0.296 & 0.436 & 0.451\\
    \cmidrule(lr){2-14}
    & \multirow{4}{*}{bi-Mamba} & \multirow{4}{*}{Attention}
    &      96 & 0.153 & 0.259 & 0.514 & 0.351 & 0.163 & 0.207 & 0.230 & 0.268 & 0.299 & 0.350\\
    & & & 192 & 0.167 & 0.266 & 0.512 & 0.348 & 0.211 & 0.252 & 0.255 & 0.287 & 0.382 & 0.401\\
    & & & 336 & 0.183 & 0.277 & 0.534 & 0.377 & \second{0.266} & \second{0.288} & 0.275 & 0.295 & 0.430 & 0.438\\
    & & & 720 & \second{0.213} & 0.311 & 0.685 & 0.441 & 0.346 & \second{0.340} & 0.284 & 0.301 & 0.433 & 0.449\\
    \cmidrule(lr){2-14}
    & \multirow{4}{*}{Attention} & \multirow{4}{*}{FFN}
    &      96 & 0.148 & 0.240 & 0.395 & 0.268 & 0.174 & 0.214 & \first{0.203} & \first{0.237} & \second{0.297} & \second{0.349}\\
    & & & 192 & 0.162 & \second{0.253} & 0.417 & 0.276 & 0.221 & 0.254 & \first{0.233} & \first{0.261} & 0.380 & 0.400\\
    & & & 336 & \second{0.178} & \first{0.269} & 0.433 & \second{0.283} & 0.278 & 0.296 & \first{0.248} & \first{0.273} & 0.428 & \second{0.432}\\
    & & & 720 & 0.225 & 0.317 & 0.467 & 0.302 & 0.358 & 0.349 & \first{0.249} & \first{0.275} & \second{0.427} & \second{0.445}\\
    \midrule
    \multirow{8}{*}{w/o} 
    & \multirow{4}{*}{bi-Mamba} & \multirow{4}{*}{w/o}
    &      96 & \second{0.141} & \second{0.238} & \first{0.380} & \first{0.259} & 0.167 & 0.214 & 0.210 & 0.250 & 0.298 & \second{0.349}\\
    & & & 192 & \second{0.160} & 0.256 & \second{0.400} & \second{0.270} & 0.217 & 0.255 & 0.245 & 0.276 & 0.381 & 0.400\\
    & & & 336 & 0.181 & 0.279 & \second{0.426} & \second{0.283} & 0.276 & 0.300 & 0.263 & 0.291 & 0.430 & 0.437\\
    & & & 720 & 0.214 & \second{0.304} & \second{0.466} & \first{0.299} & 0.353 & 0.348 & 0.268 & 0.296 & 0.433 & 0.446\\
    \cmidrule(lr){2-14}
    & \multirow{4}{*}{w/o} & \multirow{4}{*}{FFN}
    &      96 & 0.169 & 0.253 & 0.437 & 0.283 & 0.183 & 0.220 & 0.228 & 0.263 & 0.299 & 0.350\\
    & & & 192 & 0.177 & 0.261 & 0.449 & 0.287 & 0.231 & 0.262 & 0.261 & 0.283 & 0.380 & 0.399\\
    & & & 336 & 0.194 & 0.278 & 0.464 & 0.294 & 0.285 & 0.300 & 0.279 & 0.294 & \second{0.427} & 0.435\\
    & & & 720 & 0.233 & 0.311 & 0.496 & 0.313 & 0.362 & 0.350 & 0.276 & 0.291 & 0.431 & 0.449\\
    \bottomrule
  \end{tabular}
  \end{small}
  \end{threeparttable}
  }
\end{table*}

\subsection{Ablation Study}
To evaluate the efficacy of the components within S-Mamba, we conduct ablation studies by substituting or eliminating the VC and TD encoding layers. 
Specifically, the TD encoding layer is replaced with Attention, bidirectional Mamba, unidirectional Mamba, or omitted altogether (w/o). 
The choice of bidirectional Mamba (bi-Mamba), which is set to benchmark Attention, is made to facilitate global temporal information extraction. 
The rationale behind employing unidirectional Mamba is its resemblance to RNN models, so inherently possesses the capacity to preserve sequential relationships, thereby making it a suitable candidate for evaluating the impact of sequential encoding on TD. 
The VC encoding layer was replaced with an Attention mechanism or entirely removed. 
This modification is predicated on the empirical evidence from iTransformer experiments \citep{liu2023itransformer}, which demonstrate that Attention was the optimal encoder for VC. 
We do not use a unidirectional Mamba as the VC Encoding Layer, because Mamba, like RNNs, can only observe information from one direction. 
A unidirectional Mamba setting would result in the loss of half of the information, making it less effective than bidirectional Mamba or Attention in capturing global information. 

Our experimental investigations are conducted on five datasets: Electricity, Traffic, Weather, Solar Energy, and ETTh2. 
The findings from these experiments in Tab.\ref{tab:full_ab} indicate that Mamba exhibits superior performance in VC encoding, whereas the Feed-Forward Network (FFN) maintained its dominance in TD encoding. 
These findings demonstrate that S-Mamba's current framework is the most efficient. 

\begin{figure}[!htbp]
 \centering
 \includegraphics[width=0.48\textwidth]{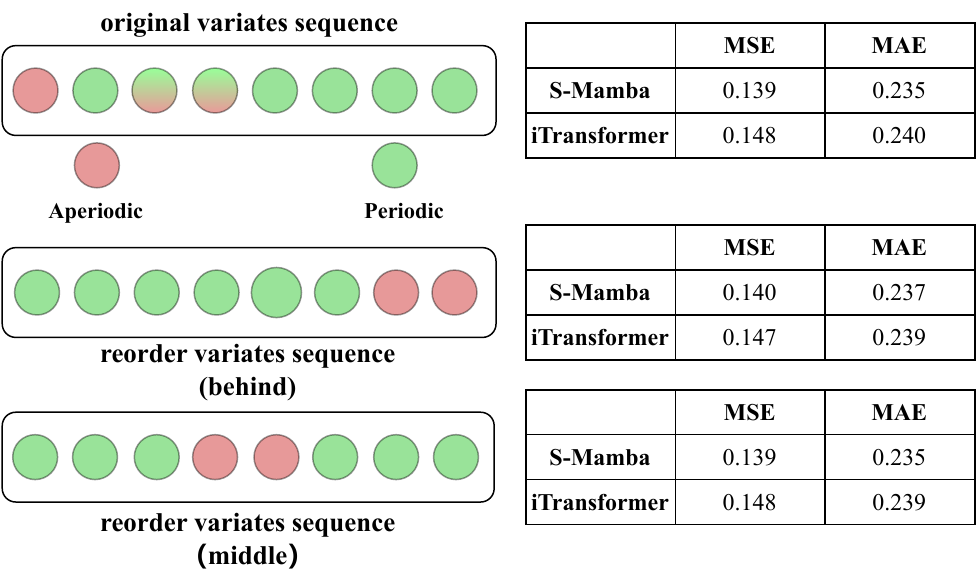}
 \caption{In the Electricity dataset, adjust the distribution of periodic and aperiodic variates. 
 The left side represents the distribution, while the right side indicates the two models' performance when lookback length $L=96$ and forecast length $T=96$.}
 \label{fig:reorder}
 \vspace{-5mm}
\end{figure}

\subsection{Can Variate Order Affect the Performance of S-Mamba?}
In S-Mamba, each variate is treated as an independent channel, so variates themselves are not inherently ordered. 
But in the Mamba VC Encoding Layer, the Mamba Block interprets the sequence like RNN, implying that it apprehends the variates as a sequence with implicit order. 
Mamba's Selective mechanism is closely linked to the Hippo matrix \citep{gu2020HiPPORM}, which causes Mamba to prioritize closer variates in sequences at initialization, leading to a bias against more distant variates. 
The initial bias towards neighboring variates may potentially impede the acquisition of a global inter-variate correlation. inspiring us to investigate the impact of the variate order on the performance of S-Mamba. 

We first use the Fourier transform \citep{bracewell1989fourier} to categorize the variates into periodic and aperiodic groups and then consider periodic variates as containing reliable information and aperiodic variates as potential noise. 
This distinction is based on the assumption that periodic variates are more likely to exhibit consistent patterns that can be learned, while aperiodic variates may contain unreliable information due to irregular fluctuations. 
Next, we decide to alter the variate order by changing the positions of these noise variates for these noise variates have the greatest impact on performance by affecting VC Encoding. 
Instead of randomly shuffling the overall variate order, it is more effective to adjust the distribution of these noisy variates. 
Subsequent trials involve repositioning the aperiodic variates towards the middle or end of the variates sequence, followed by evaluating the predictive capabilities of the models trained on these modified datasets. 
For comparative analysis, we also included experiments with the iTransformer. 

The variate distribution and corresponding model performance are illustrated in the Fig. \ref{fig:reorder}. 
We conduct this experiment only on the Electricity dataset because it requires a dataset with a large number of both periodic and aperiodic variates and Electricity is the only one that satisfies the condition. 
Our findings suggest that the S-Mamba model's performance remains largely unaffected by the perturbation of variate order. 
It implies that through adequate training, the S-Mamba can effectively mitigate the initial bias of the Hippo matrix to get accurate inter-variate correlations.

\begin{figure*}[]
 \centering
 \includegraphics[width=0.95\textwidth]{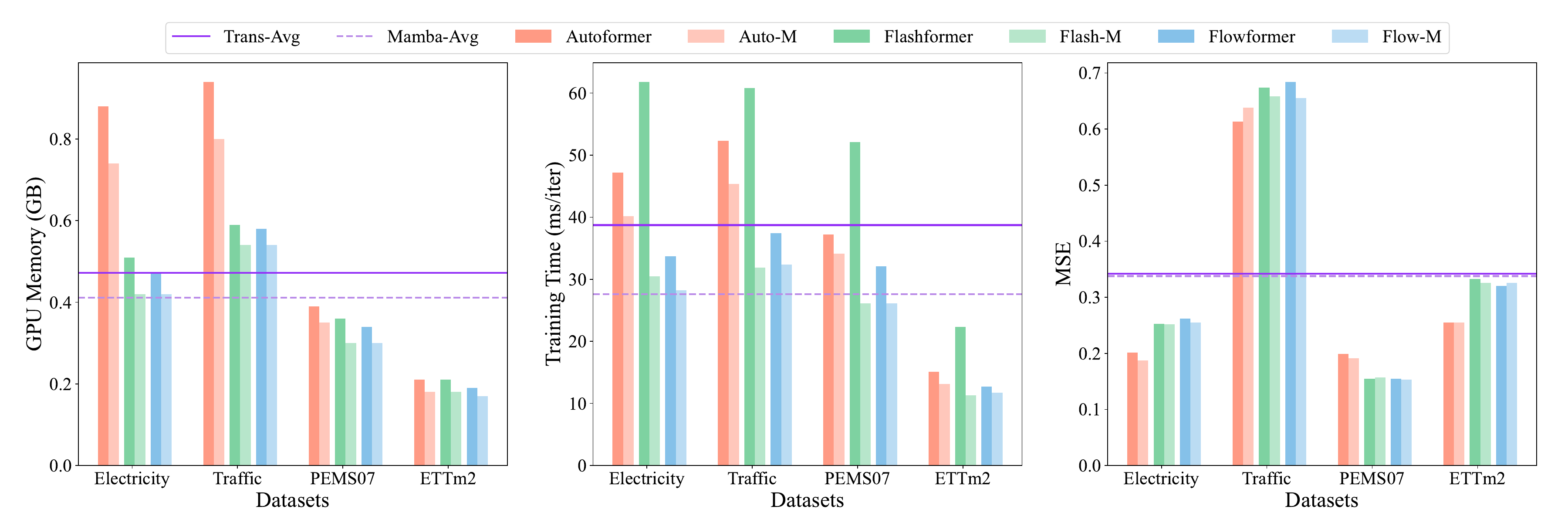}
 \caption{Allocated GPU Memory, Training Time, and MSE of Autoformer, Flashformer, Flowformer and Auto-M, Flash-M, and Flow-M on four datasets. The lookback length $L=96$ and the forecast length $T=12$ for PEMS07, $T=96$ for other datasets.
 The purple horizontal line represents the average performance of "Transformer models", and the purple dotted line represents the average performance of "Mamba models".}
 \label{fig:G_T_M}
 \vspace{-5mm}
\end{figure*}

\subsection{Can Mamba Outperform Advanced Transformers?}
Beyond the foundational Transformer architecture, some advanced Transformers have been introduced, predominantly focusing on the augmentation of the self-attention mechanism. 
We aim to determine whether Mamba can still maintain an advantage over these advanced Transformers. 
To the end, we conduct a comparative experiment in which we directly replace the Encoder layer of three advanced self-attention mechanism in three Transformer: Autoformer \citep{wu2021autoformer}, Flashformer \citep{dao2022flashattention} and Flowformer \citep{wu2022flowformer} with a unidirectional Mamba (uni-Mamba) for TSF tasks to get Auto-M, Flash-M and Flow-M to compare their performance. 
The reason behind using a uni-Mamba is that the Encoder layer of these three Transformers handles Temporal Dependency (TD), which is inherently ordered. 
Therefore, a uni-Mamba is more suitable than a bidirectional Mamba, for apprehending the sequential nature of TD. 

We compare the GPU Memory, training time, and MSE of three advanced Transformer models and their Mamba Encoder counterparts on Electricity, Traffic, PEMS07 and ETTm1 as Fig. \ref{fig:G_T_M}. 
The findings indicate that employing Mamba as the Encoder directly resulted in reduced GPU usage and training time consumption while achieving slightly improved overall performance. 
It means that Mamba can still maintain its advantage compared to these advanced self-attention mechanisms or, in other words, these advanced Transformers. 

\begin{figure*}[]
 \centering
 \includegraphics[width=0.98\textwidth]{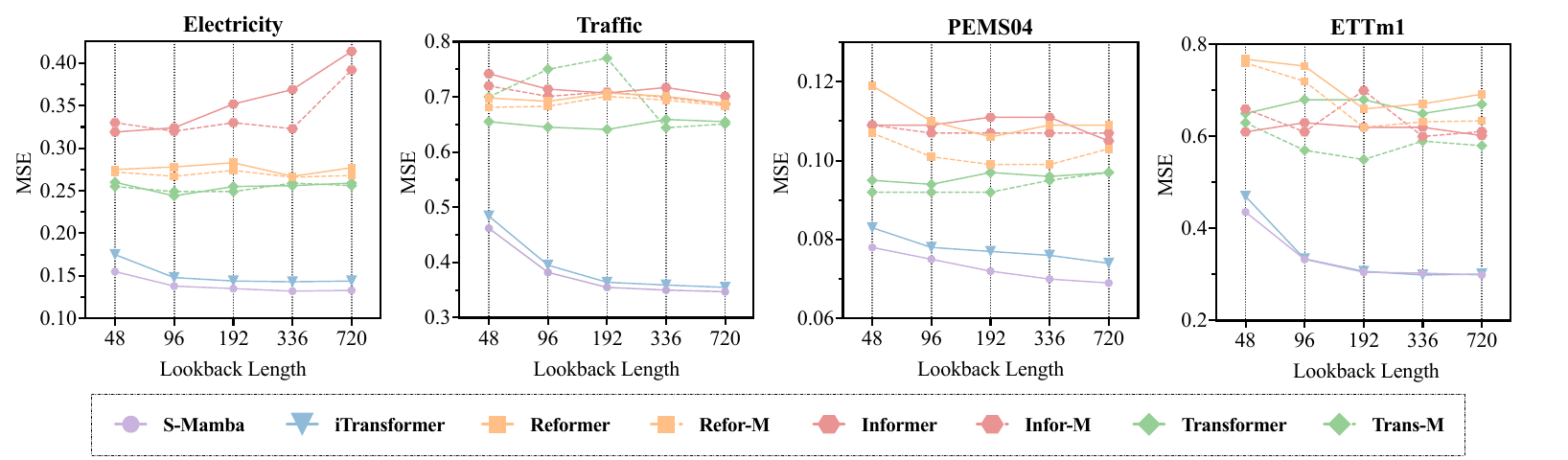}
 \caption{Forecasting performance on four datasets with the lookback length $L \in \{ 96,192,336,720\}$ while the forecast length $T = 12$ for PEMS04 and $T = 96$ for other datasets.}
 \label{fig:length}
 \vspace{-3mm}
\end{figure*}

\subsection{Can Mamba Help Benefit from Increasing Lookback Length?}
Prior research shows that Transformer-based models' performance does not consistently improve with increasing lookback sequence length $L$, which is somewhat unexpected. 
A plausible explanation is that the temporal sequence relationship is overlooked under the self-attention mechanism, as it disregards the sequential order, and in some instances, even inverts it. 
Mamba, resembling a Recurrent Neural Network \citep{medsker2001recurrent}, concentrates on the preceding window during information extraction, thereby preserving certain sequential attributes. 
It prompts an exploration of Mamba's potential effectiveness in temporal sequence information fusion, aiming to address the issue of diminishing or stagnant performance with increasing lookback length.  
Consequently, we add an additional Mamba block between the encoder Layer and decoder layer of Transformers-based models. 
The role of the Mamba Block is to add a layer of time sequence dependence from the information output by the encoder layer, to add some information similar to position embedding before the decoder layer processes it. 
We experiment with Reformer \citep{wu2021autoformer}, Informer \cite{zhou2021informer}, and Transformer \citep{vaswani2017attention} to get Refor-M, Infor-M, and Trans-M, and evaluate their performance with varying lookback lengths. 
We also test the performance of S-Mamba and iTransformer as the lookback length increases. 
The experiment is conducted on four datasets: Electricity, Traffic, PEMS04 and ETTm1. 
The results are in Fig. \ref{fig:length}, from which we can observe four results. 
(1) S-Mamba and iTransformer can enhance their performance as the input lengthens, but we believe it is not solely due to the Mamba Block or Transformer Block, but rather to the FFN TD Encoding Layer they both possess. 
(2) S-Mamba consistently outperforms iTransformer, primarily due to the superior performance of S-Mamba's Mamba VC Encoding layer compared to iTransformer's Transformer VC Encoding layer.
(3) After incorporating the Mamba Block between the Encoder and Decoder layer, performance enhancements are typically observed in the original model across the four datasets.  
(4) Despite these variants' performance gains sometimes, they do not achieve optimization with longer lookback lengths. 
It is consistent with the findings of \citet{zeng2023dlinear}, which also suggests that encoding temporal sequence information into the model beforehand does not resolve the issue. 

\begin{figure}[tbp]
 \centering
 \includegraphics[width=0.45\textwidth]{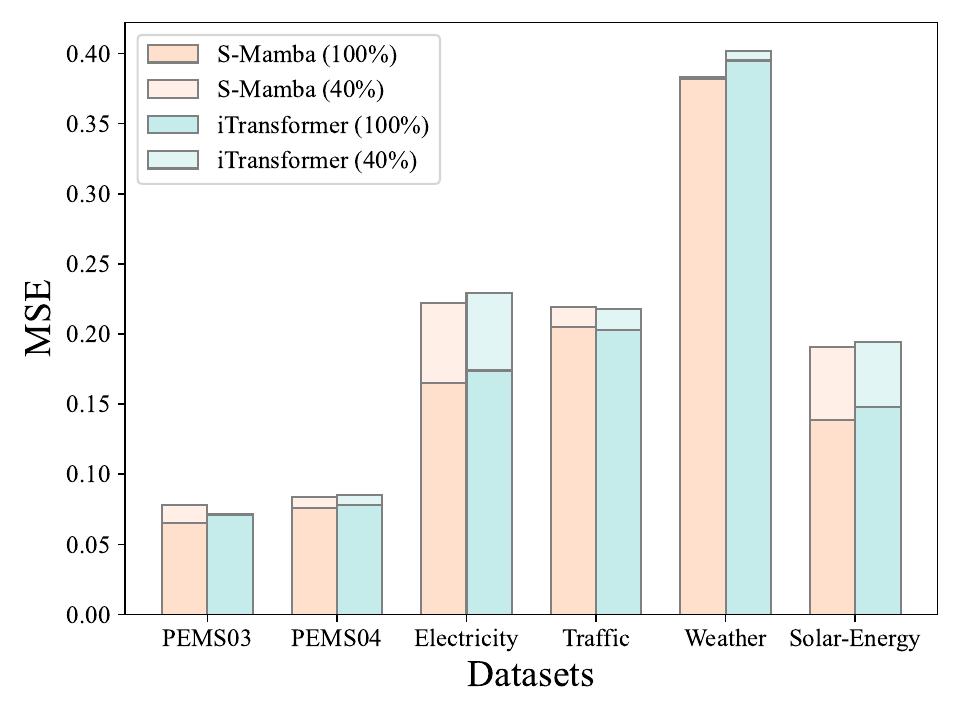}
 \caption{Forecasting performance comparison between S-Mamba and iTransformer trained on 100\% variates with on 40\% variates. 
 The lookback length $L = 96$ for all datasets. 
 For the PEMS dataset, the output length $T = 12$, while for the other datasets $T = 96$.}
 \label{fig:gen}
 \vspace{-5mm}
\end{figure}
\subsection{Is Mamba Generalizable in TSF?}
The emergence of pretrained models \citep{devlin2018bert} and large language models \citep{chang2023survey} based on the Transformer architecture has underscored the Transformer's ability to discern similar patterns across diverse data, highlighting its generalization capabilities. 
In the context of TSF, it is observed that some variates exhibit a similar pattern of differences, so the generalization potential of the Transformer for sequence data may also take effect on TSF tasks. 
In this vein, iTransformer \citep{liu2023itransformer} conducts a pivotal experiment. 
The study involves masking a majority of the variates in a dataset and training the model on a limited subset of variates. 
Subsequently, the model was tasked with forecasting all variates, including those previously unseen, based on the learned information from the few varieties. 
The results show that Transformer can use its generalization ability to make accurate predictions for unseen variates in TSF tasks. 
Building on it, we seek to evaluate the generalization capabilities of Mamba in TSF tasks. 
An experiment is proposed wherein the S-Mamba are trained on merely 40\% of the variates in the PEMS03, PEMS04, Electricity, Weather, Traffic, and Solar datasets. 
We selected these datasets for testing because they contain a large number of variates, which makes it fair to evaluate the models' generalization ability. 
Then they are employed to predict 100\% variates, and the results are subjected to statistical analysis.
The outcomes of this investigation in \ref{fig:gen} reveal that S-Mamba exhibits generalization potential on the six datasets, which proves their generalizability in TSF tasks. 

\section{Conclusion}
Transformer-based models have consistently exhibited outstanding performance in the field of time series forecasting (TSF), while Mamba has recently gained popularity, and has been shown to surpass the Transformer in various domains by delivering superior performance while reducing memory and computational overhead. 
Motivated by these advancements, we seek to investigate the potential of Mamba-based models in the TSF domain, to uncover new research avenues for this field. 
To this end, we introduce a Mamba-based model for TSF, Simple-Mamba (S-Mamba). 
It transfers the task of inter-variate correlation (VC) encoding from the Transformer architecture to a bidirectional Mamba block and uses a Feed-Forward Network to extract Temporal Dependencies (TD). 
We compare S-Mamba with nine representative and state-of-the-art models on thirteen public datasets including Traffic, Weather, Electricity, and Energy forecasting tasks. 
The results indicate that S-Mamba requires low computational overhead and achieves leading performance. 
The advantage is primarily attributed to the bidirectional Mamba (bi-Mamba) block within the Mamba VC Encoding Layer, which offers an enhanced understanding of VC at a lower overhead compared to the Transformer. 
Furthermore, we conduct extensive experiments to prove Mamba possesses robust capabilities in TSF tasks. 
We demonstrate that the Mamba maintains the same stability as the Transformer in extracting VC and still can offer advantages over advanced Transformer architectures. 
Transformer architectures can see performance gains by simply integrating or substituting with Mamba blocks. 
Additionally, Mamba exhibits comparable generalization capabilities to the Transformer. 
In a word, Mamba exhibits remarkable potential to outperform the Transformer in the TSF tasks. 

\section{Future Work}
As the number of variates grows, global inter-variate correlations (VC) become increasingly valuable and the extraction of them becomes more difficult and consumes more computational resources. 
Mamba excels at detecting long-range dependencies and controlling the escalation of computational demands, thus equipping it to meet the challenges outlined. 
In real-life scenarios where resources are limited, compared with Transformer, Mamba is capable of processing more variates information simultaneously and delivering more accurate predictions.  
For example, in traffic forecasting, Mamba can rapidly assess traffic flows at more intersections, and in hydrological forecasting, it can provide insights into conditions across more tributaries. 
Looking forward, Mamba-based models are expected to be applicable to a broader spectrum of time series prediction tasks that involve processing extensive variate data. 

Pretrained models utilizing the Transformer architecture capitalize on its robust generalization capabilities, achieving notable success in TSF. 
These models demonstrate effectiveness across various tasks through fine-tuning. 
Our experimental results indicate that Mamba matches the Transformer in both generalization and stability, suggesting that the development of a Mamba-based pre-training model for TSF tasks could be a fruitful direction to explore. 
% Use the style of numbering in square brackets.
% If nothing is used, default style will be taken.
%\begin{enumerate}[a)]
%\item 
%\item 
%\item 
%\end{enumerate}  

%\begin{itemize}
%\item 
%\item 
%\item 
%\end{itemize}  

%\begin{description}
%\item[]
%\item[] 
%\item[] 
%\end{description}  

% Figure
% \begin{figure}[<options>]
% 	\centering
% 		\includegraphics[<options>]{}
% 	  \caption{}\label{fig1}
% \end{figure}

% \begin{table}[<options>]
% \begin{tabular*}{\tblwidth}{@{}LL@{}}
% \toprule
% \midrule
%  & \\
%  & \\
%  & \\
%  & \\
% \bottomrule
% \end{tabular*}
% \end{table}

% Uncomment and use as the case may be
%\begin{theorem} 
%\end{theorem}

% Uncomment and use as the case may be
%\begin{lemma} 
%\end{lemma}



\bibliographystyle{cas-model2-names}

\bibliography{refs}

% % Here goes the biography details.

% % Here goes the biography details.

\end{document}